%% file: main.tex
\documentclass[11pt]{article}

\usepackage[preprint]{acl}

\usepackage{kotex}
\usepackage{times}
\usepackage{latexsym}
\usepackage{amsmath}
\usepackage{amssymb}
\usepackage{colortbl}
\usepackage[utf8]{inputenc}
\usepackage[T1]{fontenc}
\usepackage{hyperref}
\usepackage{url}
\usepackage{booktabs}
\usepackage{amsfonts}
\usepackage{nicefrac}
\usepackage{microtype}
\usepackage{xcolor}
\usepackage{algorithm}
\usepackage{algpseudocodex}
\usepackage{tcolorbox}
\usepackage{multirow}
\usepackage{subfig}
\usepackage{graphicx}
\usepackage{placeins}
\usepackage{dblfloatfix}
\usepackage{float}
\usepackage{inconsolata}

\newcommand{\ie}{\textit{i}.\textit{e}., }
\newcommand{\eg}{\textit{e}.\textit{g}., }

\input{figure_table}

\title{Colla-Q: Toward Collaborative Experts in MoE Quantization \\ via Minimax Precision Balancing}

\author{Eunju Shin \\
  Ajou University, South Korea \\
  \texttt{eunjushin@ajou.ac.kr} \\\And
  Jongbin Ryu\thanks{Corresponding author.} \\
  Ajou University, South Korea \\
  \texttt{jongbinryu@ajou.ac.kr} \\
}

\begin{document}
\maketitle

\begin{abstract}

In this paper, we present a Mixture-of-Experts (MoE) quantization method based on activation entropy.
Although quantization reduces memory and computational costs, it can substantially degrade performance.
In particular, performance decline is pronounced in quantized MoE models, where individual experts have a small number of parameters that are sensitive to low-bit representation.
Considering that MoE operates as an ensemble model with collaborative contributions from routed experts, a significant performance decline of a particular expert due to quantization can harm model performance.
Therefore, we propose \textbf{Colla-Q}, a bit-allocation framework to maintain balanced performance across experts through an activation-entropy-based bit-width allocation algorithm.
This approach encourages each expert to operate collaboratively in the quantized model, thereby 1) improving the overall MoE performance and 2) reducing the dependence on the calibration dataset.
Since uniformly adjusting each expert's performance facilitates robustness and stability of the MoE model, the proposed MoE quantization method can
generalize more consistently across different calibration datasets.
Our code is available at: \url{https://github.com/mmai-laboratory/Colla_Q}
\end{abstract}

\section{Introduction}
In recent large language models (LLMs), MoE has been widely adopted as a sparse architecture to reduce computational overhead~\citep{jiang2024mixtral,muennighoff2025olmoe}.
It reduces overhead by routing input tokens to relevant experts during inference~\citep{shazeer2017outrageously}.
However, despite this efficiency, MoE models require substantial memory to load all expert parameters, including those that are not routed.
This leads to excessive memory requirements as the model size increases.
Therefore, mixed-precision quantization is an effective method to address this limitation of MoE.
However, mixed-precision quantization methods developed for dense LLMs are not suitable for MoE due to the unique structure of MoE architectures.

\introfigureTwo
\introfigure

Unlike dense LLMs, MoE models consist of individual experts with relatively few parameters, each tending to specialize in different token contexts.
As a result, experts are highly sensitive to low-bit quantization, and some experts experience significant performance degradation~\citep{dong2019hawq}.
This degradation is non-uniform across experts, which can amplify performance differences and impair the quality of the aggregated MoE output.
This phenomenon is consistent with an ensemble perspective: previous
studies~\citep{caruana2004ensemble,an2023selective,yao2025determine}
show that a single weak learner can hinder collaborative expert contributions and, in turn, degrade the overall performance of the aggregated model.
This perspective motivates our view of MoE architectures, as they aggregate the contributions of routed experts in an ensemble-like manner.
We provide experimental results supporting these observations in Figure~\ref{fig:introfiguretwo}, which demonstrates the relationship between performance balancing across experts and overall model performance.
Accordingly, this observation motivates assigning more appropriate bit-widths to weak experts in MoE quantization to reduce further performance degradation.

Previous studies~\citep{huang2025mixture,li2024quantmoe,duanmu2025mxmoe} have made similar efforts to optimize MoE bit-width allocation. They allocate different bit-widths to experts based on routing frequency. \citet{huang2025mixture} and \citet{li2024quantmoe} measure each expert’s importance by counting how often tokens are routed to it on a calibration dataset. Frequently routed experts are deemed more important due to higher utilization and therefore receive more bits. \citet{duanmu2025mxmoe} also uses per-expert routing results to guide bit-width allocation.
 Although these methods are effective for MoE quantization, they do not directly consider preserving the performance of weak experts. As shown in Figures~\ref{fig:compare_balance} and~\ref{fig:introfigure}, allocating more bits to weak experts suggests that promoting similar expert performance after quantization helps preserve the aggregated MoE block output. Moreover, their estimates of expert importance depend on the calibration distribution, as shown in Figure~\ref{fig:mixtralcalb}.

Therefore, we emphasize that maintaining balanced expert performance is important, as it supports collaborative expert contributions and ultimately improves overall model performance.
The key question at this point is how to estimate expert performance for bit-width allocation.
Since individual experts do not directly generate tokens,
we need a label-free way to estimate expert performance from output activations.
To this end, we propose an activation-entropy metric that estimates expert performance through the lens of activation variance, grounded in differential entropy theory. Based on this metric, we further introduce a minimax precision balancing algorithm to preserve balanced expert performance. We refer to this overall framework as \textbf{Colla-Q}.

Colla-Q also shows better generalization across calibration datasets. For practical quantization, this ability to generalize across calibration datasets is important because the inference distribution is often unknown in advance. However, routing frequency and weights used for bit-width allocation in MoE quantization tend to reflect properties of the calibration domain more strongly, as shown in Table~\ref{tab:calb_sim_avg}. For example, when calibrating a model on a dataset with math problems, these routing-based metrics can place greater emphasis on experts specialized in numerical reasoning. By contrast, as shown in Figure~\ref{fig:mixtralcalb}, our method is less sensitive to the calibration domain because it allocates bit-widths based on expert performance estimated from output activations. As a result, this yields more consistent bit-width assignments and smaller performance variation across calibration datasets.

\section{Related Work}
\subsection{Quantization for LLMs with Activation}
Post-training quantization (PTQ) is a method that quantizes pre-trained models without the need for additional training, effectively reducing the memory footprint of large language models (LLMs).
GPTQ~\citep{frantar2022gptq} is a representative PTQ method that enables low-bit quantization via layer-wise optimization.
It performs layer-wise optimization to minimize quantization error for each layer using calibration data.
Meanwhile, recent methods such as LLM.int8()~\citep{dettmers2022gpt3}, OWQ~\citep{lee2024owq}, and AWQ~\citep{lin2024awq} have shown that performance degradation can be reduced using activation statistics.
In particular, AWQ identifies important channels based on activation statistics and scales their weights to reduce performance degradation during quantization. These results suggest that activation-derived signals are useful for enhancing the effectiveness of weight quantization.

\subsection{Quantization for MoE-LLMs: From Allocation to Precision Balancing}
Quantization is widely applied to MoE models to reduce the memory footprint while mitigating performance degradation.
Across several studies~\citep{li2024quantmoe,huang2025mixture,duanmu2025mxmoe}, mixed precision is applied to different MoE components, such as MoE blocks, experts,  and linear layers within each expert.
QuantMoE-Bench~\citep{li2024quantmoe} presents a structure-aware strategy and shows that mixed precision outperforms uniform bit-width under the same budget.
Building on this strategy, PMQ~\citep{huang2025mixture} allocates per-expert bit-width using routing statistics from a calibration dataset, such as routing frequency and weight.
MxMoE~\citep{duanmu2025mxmoe} considers linear-layer quantization sensitivity, expert activation patterns, and hardware constraints to allocate bit-width.
Overall, prior work has designed mixed-precision bit-width for different MoE components.
Most of these approaches determine expert importance using routing statistics from a calibration dataset, implicitly assuming that more frequently activated experts are more critical.
In contrast, rather than allocate bit-width based on routing statistics, our minimax precision balancing approach actively equilibrates expert performance to maintain the integrity of the overall ensemble.

\section{Method}
Previous studies on ensemble models~\citep{caruana2004ensemble,an2023selective,yao2025determine}
have demonstrated that when a single weak learner performs at an exceptionally low level, the performance of the aggregated ensemble model drops significantly.
To address this performance drop in the quantized MoE model,
we adopt an ensemble perspective as the motivation for our approach.
Building on this perspective,
we present \textbf{Colla-Q}, a bit-allocation framework with two components: 1) an activation-entropy metric for expert performance estimation and 2) a minimax precision balancing algorithm for bit allocation.
\subsection{MoE from Ensemble Perspective}
\label{sec:moeensemble}
Formally, the output of a MoE block can be understood as a dynamic ensemble output of $k$ routed experts.
Given an input $\mathbf{x}$, the layer output $\hat{\mathbf{y}}$ is derived from the weighted ensemble of the expert outputs as:
\begin{equation}
\hat{\mathbf{y}} = \sum_{e \in \hat{E}} g_e(\mathbf{x}) \cdot e(\mathbf{x}),\
\end{equation}
where $\hat{E}$ is the set of routed experts, $e(\mathbf{x})$ refers to the expert output, and $g_e(\mathbf{x})$ indicates the corresponding routing weight.
Let $R_e$ represent the expected error of an expert, which quantifies its individual performance deficiency.
Assuming the errors of distinct experts are uncorrelated, we can approximate the global error $\mathcal{E}_{ens}$ of the aggregated ensemble by the weighted sum of individual errors:
\begin{equation}
\mathcal{E}_{ens} \approx \sum_{e \in \hat{E}} (\bar{g}_e(\mathbf{x}))^2 R_e.\
\end{equation}

In order to enhance the stability and robustness of the model, our goal is to minimize the global ensemble error $\mathcal{E}_{ens}$.
Considering a fixed total amount of expert errors (\ie $\sum_{e\in\hat{E}} R_e = \mathcal{A}$), for simplicity, we assume approximately uniform routing weights in expectation over many tokens, $\bar{g}_e \triangleq \mathbb{E}_{\mathbf{x}}[g_e(\mathbf{x})] \approx \frac{1}{|\hat{E}|}$ as the load-balancing objective,
which encourages balanced expert utilization on average over many tokens (shown in Appendix~\ref{sec:a1_routingweights}).
Under these assumptions, we can structure the optimization problem as minimizing the following convex objective:
\begin{equation}
\min_{\{R_e\}} \sum_{e \in \hat{E}} R_e^2 \quad \text{s.t.} \quad \sum_{e \in \hat{E}} R_e = \mathcal{A}.\
\end{equation}

Since the objective function is strictly convex, we obtain the global minimum only when the errors across all routed experts are identical:
\begin{equation}
R_{e_i} \approx R_{e_j} \approx  \frac{\mathcal{A}}{|{\hat{E}}|} \quad \forall e_i, e_j \in \hat{E}.\
\end{equation}

This analysis suggests that differences in expected errors among experts can increase the lower bound of the global ensemble error.
Therefore, according to an ensemble perspective, we are motivated to encourage similar performance across all experts to strengthen the robustness of MoE.

\subsection{Activation-Entropy of Experts}
\label{sec:activationentropy}
We introduce the activation-entropy metric in Colla-Q, which estimates expert performance without using token labels.
Typically, language-model performance is measured by cross-entropy, which compares generated tokens with token labels.
However, this approach is not feasible in our method, because experts do not directly generate tokens; instead, we evaluate each expert using its continuous-valued output activations (\ie expert FFN outputs), without access to token labels.
Therefore, we employ the principle of differential entropy~\citep{cover1999elements,malinin2018predictive} that quantifies uncertainty to derive the entropy of experts, as follows:

\begin{equation}
\label{eq:diff_entropy}
H(x) = \frac{1}{2} \ln(2\pi e \sigma^2),
\end{equation}
where $H(x)$ is the differential entropy, $\pi$ and $e$ denote the constant values, and $\sigma^2$ represents the variance of $x$.
Eq.~\ref{eq:diff_entropy} holds when $x$ follows the Gaussian approximation $p(x) \sim \mathcal{N}(\mu, \sigma^2)$, and
the activations generated by each expert are reasonably well approximated by this distribution (shown in Appendix~\ref{sec:a2_gaussian}).
Therefore, by applying Eq.~\ref{eq:diff_entropy}, we can effectively express expert activation-entropy in terms of activation variance.
Based on this, we define the activation-entropy-based proxy $\rho$ for expert performance by the following ratio:

\begin{equation}
\label{eq:ratio_metric}
\rho = \frac{\exp(2H_{within})}{\exp(2H_{total})} \propto \frac{\sigma_{within}^2}{\sigma_{total}^2},
\end{equation}
where lower values of $\rho$ are associated with better expert performance.
Here $H_{within}$ and $H_{total}$ are the within-channel and total entropies of activations. By Eq.~\ref{eq:diff_entropy}, entropy is determined by variance, so we compute the corresponding variances as:

\begin{equation}\label{eq:variance}
\begin{aligned}
\sigma_{\mathrm{within}}^2
=\frac{1}{|\mathcal{T}_e||\mathcal{C}|}\sum_{t\in\mathcal{T}_e}\sum_{c\in\mathcal{C}} \big(A_{t,c}-\mu_c\big)^2, \\
\sigma_{\mathrm{total}}^2
=\frac{1}{|\mathcal{T}_e||\mathcal{C}|}\sum_{t\in\mathcal{T}_e}\sum_{c\in\mathcal{C}} \big(A_{t,c}-\bar{\mu}\big)^2,
\end{aligned}
\end{equation}
where \(\mathcal{T}_e\) is a token set routed to expert \(e\), \(\mathcal{C}\) denotes channels of the activations,
and  \(A_{t,c}\) is the activation at token $t$ and channel $c$.
The channel-wise and global averages are denoted as \(\mu_c = \frac{1}{|\mathcal{T}_e|}\sum_{t\in\mathcal{T}_e} A_{t,c}\) \text{and} \(\bar{\mu} = \frac{1}{|\mathcal{T}_e||\mathcal{C}|}\sum_{t\in\mathcal{T}_e}\sum_{c\in\mathcal{C}} A_{t,c}\).

The denominator in Eq.~\ref{eq:ratio_metric}, total entropy and variance, signifies the expert’s capacity to represent activations in its global activation space.
It measures the dynamic range of activations across neurons, reflecting the expert’s capacity to represent diverse features for the tokens routed to the expert.
The numerator in Eq.~\ref{eq:ratio_metric}, the within-channel entropy and variance, quantifies the uncertainty across channels.
Each channel of the activations is hypothesized to act as a dedicated feature encoder for specific input patterns, and thus an expert is expected to produce predictable responses.
A high within-channel variance suggests that an expert is less predictable, reflecting high uncertainty.

A related intuition also appears in random forest~\citep{breiman2001random}, which explains the generalization ability of the ensemble model as the strength and correlation of decision trees.
In random forests, the generalization ability is quantified by the ratio of the strength, which measures the uncertainty of decision trees, and the correlation, which evaluates their representation capacity.
Therefore, the proposed ratio \(\rho\) can effectively estimate expert performance. Lower values of \(\rho\) suggest that an expert captures its unique channel-wise patterns more consistently while achieving high representational capacity.
Additional empirical analysis of activation entropy is provided in Appendix~\ref{sec:santiycheck}.

\subsection{Minimax precision balancing}
We introduce the second component of Colla-Q, a minimax precision balancing algorithm that uses the proposed activation-entropy metric to allocate bit-widths across experts. We begin by revisiting the objective function of previous MoE quantization methods~\citep{huang2025mixture} as follows:
\begin{equation}
\begin{aligned}
\label{eq:obj_previous}
\min_{\mathbf{b}} \sum_{i=1}^{N} \mathcal{L}({b_{e_i}}) \quad \text{s.t.} \quad &\sum_{i=1}^{N} b_{e_i} \le B_{total}\\
\quad &1 \le b_{e_i} \le B_{max}
\end{aligned}
\end{equation}
where $E$ denotes the expert set of a single MoE block, $b_e$ is a bit-width of expert $e$, and $B_{total}$ stands for the total bit-width assigned for the MoE block.
$\mathcal{L}(b_e)$ is an objective function that should be minimized during the mixed-precision quantization process.
Therefore, according to Eq.~\ref{eq:obj_previous}, previous mixed-precision quantization methods minimize the averaged loss of the objective functions across all experts.
However, our approach targets the minimization of the objective function of the worst-performing expert, as follows:
\begin{equation}
\label{eq:obj_ours}
\mathbf{b}^* = \underset{\mathbf{b}}{\arg \min} \left( \max_{i=1}^{N} \mathcal{L}(b_{e_i}) \right)
\end{equation}
Under this problem definition, we aim to allocate optimal bit-widths $\mathbf{b}^* = \{b_{e_1}, b_{e_2}, \dots, b_{e_N}\}$ to $N$ experts within a MoE block, such that expert performance is balanced after mixed-precision quantization.
To support collaboration among experts, we prioritize experts based on their full-precision performance estimated using the activation-entropy metric in the objective function as:

\begin{equation}
\label{eq:our_obj}
\mathcal{L}(b_{e_i})
= \rho_{e_i}\times
\underbrace{\|\,e_i(\mathbf{x})-\tilde{e}_i(\mathbf{x}; b_{e_i})\,\|_2^2}_{\text{Quantization error}},
\end{equation}
where $e_i(\mathbf{x})$ denotes the full-precision output of expert $e_i$, and
$\tilde{e}_i(\mathbf{x}; b_{e_i})$ denotes the output after applying GPTQ quantization to $e_i$ with bit-width $b_{e_i}$.
The quantization error is also considered to capture further degradation of weak experts after quantization, which can affect the aggregated MoE output.
Based on this objective function, we formalize this objective through the minimax precision balancing algorithm as shown in Algorithm~\ref{alg:ours}.
By recursively assigning additional bits to the worst-performing expert, our minimax precision balancing algorithm forces the ensemble into a balanced state.

\oursalgorithmtwo
As shown in Algorithm 1, after initialization, we recursively assign one additional bit to the worst-performing expert. This procedure emphasizes that the performance of a specific expert is not significantly degraded, unlike previous bit-width allocation approaches, which optimize all experts equally to maximize average performance. In other words, we posit that balancing expert performance ultimately benefits the MoE, even if it does not maximize the average performance. This approach, which emphasizes collaboration among experts, improves MoE performance while reducing calibration-specific overfitting in bit allocation, thereby maintaining generalizability.

\section{Experiments}
\label{sec:experiments}
\Mixtraltable
\Othersmodeltalbe
\mmlufiveshot

\Mixtralcalb

\subsection{Settings}
\textbf{Model and datasets.}
We conduct extensive experiments on three MoE models: Mixtral 8×7B~\citep{jiang2024mixtral}, DeepSeek-MoE-16B-Base~\citep{dai2024deepseekmoe}, and Phi3.5-MoE~\citep{abdin2024phi}.
The architectural details and parameter scales of each model are provided in Appendix~\ref{sec:appendiximple}.
We evaluate our method using EleutherAI’s LM Evaluation Harness~\citep{eval-harness} on eight zero-shot benchmarks: PIQA~\citep{bisk2020piqa}, ARC-Easy and ARC-Challenge~\citep{clark2018think}, BoolQ~\citep{clark2019boolq}, HellaSwag~\citep{zellers2019hellaswag}, WinoGrande~\citep{sakaguchi2021winogrande}, MathQA~\citep{amini-etal-2019-mathqa}, and MMLU~\citep{hendrycks2021ethics}.
In addition, we use four calibration datasets: C4~\citep{raffel2020exploring}, MathQA, BoolQ, and the French subset of Lambada\_mul\_fr~\citep{paperno-EtAl:2016:P16-1}.

\noindent \textbf{Experimental Setup.}
We determine the bit-width allocation for our method using 128 sequences randomly sampled from C4. For the generalizability analysis, we repeat bit-width allocation using each calibration dataset (C4, Math, French, and QA). We then evaluate the proposed method against state-of-the-art MoE quantization methods.
Across all experiments, we use GPTQ for weight quantization, with 128 sequences sampled from Wikitext2~\citep{merity2017pointer} as the GPTQ calibration dataset.
Additional implementation details for our method and the baselines are provided in Appendix~\ref{sec:appendiximple}.

\subsection{Experimental Results}
\label{sec:comparebaseline}
\noindent \textbf{SOTA comparison.}
Table~\ref{tab:mixtral87b_main} compares our method with Uniform GPTQ\footnote{Uniform\_2 allocates 2-bit uniformly to all experts in a single-precision manner.} and state-of-the-art (SOTA) mixed-precision MoE quantization methods on Mixtral 8$\times$7B.
The results show that our method performs favorably across most average bit-width settings.
A similar trend is observed for DeepSeek-16B-Base and Phi3.5-MoE in Table~\ref{tab:deepsk-phi}, where our method consistently performs well.
Our method exhibits almost no average performance drop at an average bit-width of 2.54 bits on Mixtral 8$\times$7B and DeepSeek-16B-Base. We verify in-context performance using MMLU 5-shot on Mixtral 8$\times$7B (Table~\ref{tab:mmlufiveshot}), complementing the zero-shot comparisons. The conclusions remain unchanged under the 2.54-bit setting, indicating that our bit allocation remains effective in the few-shot setting.

\noindent \textbf{Generalizability.}
We analyze the generalizability of quantization methods across different calibration datasets.
As shown in Table~\ref{tab:calb_sim_avg}, PMQ exhibits lower cosine similarity between routing-based factor vectors (routing frequency and routing weight) computed from different calibration datasets.
This result indicates that PMQ’s routing-based metrics lead to substantially different bit-width configurations depending on the calibration dataset, as the routing-based metric can change with the calibration distribution.
In contrast, our method maintains high cosine similarity, resulting in consistent bit-width configurations across calibration datasets.
Figure~\ref{fig:mixtralcalb} also illustrates this property by comparing quantization performance across MoE models under different calibration datasets.
The performance of PMQ deteriorates considerably when the calibration dataset differs from the test dataset; however, our method shows favorable performance across most calibration datasets.
These results support the generalizability of our method across calibration datasets in the quantization process.
\calbfactorcos

\quantlosspmqours
\totalalgorithmtable
\subsection{Ablation Study}

\label{sec:ablationstudy}
We provide experimental results for the two main components of our method: the proposed activation-entropy metric and the minimax precision balancing algorithm.
Table~\ref{tab:quantlosspmqours} demonstrates that the activation-entropy metric performs favorably compared to the routing-based metric used in prior quantization methods.
We conduct experiments in which implementation details are identical, differing only in the metric used in the objective function to determine the bit-width configuration.
Under this ablation setting, our activation-entropy metric consistently achieves strong results.
We validate the effectiveness of our minimax precision balancing algorithm in Table~\ref{tab:totalalrorithmtable}, where our method compares favorably with PMQ’s allocation algorithm.
Considering our method is designed to prioritize activation-entropy rather than a routing-based metric, this result aligns with the ensemble perspective of MoE.
Since we enhance the worst-performing expert by recursively adding bits, we can reduce the error of the aggregated output of multiple experts.
Additional results isolating the effects of the metric and allocation strategy, together with further algorithmic analysis, are provided in Appendix~\ref{sec:additional_ablation}.

We further examine the role of activation entropy by comparing quantization error alone with activation-entropy-weighted quantization error in Table~\ref{tab:quantlosstable}. Activation entropy serves as a label-free proxy for estimating expert performance from full-precision activations, while the expert-output quantization error captures quantization sensitivity. The combined objective consistently improves performance across all average bit-widths, with a larger gain in the extreme low-bit setting, indicating the benefit of jointly considering weak experts and quantization sensitivity. Also, we analyze our design choice in Table~\ref{tab:linearexpert}, where expert-wise mixed-precision allocation yields better performance across all average bit-width settings, supporting our design choice. Full results for Tables~\ref{tab:quantlosstable} and~\ref{tab:linearexpert} can be found in Appendix~\ref{sec:complete_ablation}.

\quantlosstable
\linearexpertcomparison
\ablationcompute
Additionally, we measure the expert-computation efficiency of Colla-Q using low-bit expert GEMM kernels and compare it with FP16 under the same routing workload. As shown in Figure~\ref{fig:ablationcompute}, Colla-Q achieves higher TFLOPS and a $1.39\times$--$1.48\times$ expert-computation speedup across the evaluated average bit-widths. These results suggest that, beyond reducing memory usage while preserving performance, Colla-Q can also provide practical computation gains with low-bit expert GEMM kernels. This analysis focuses on expert computation rather than end-to-end inference, and detailed experimental settings are provided in Appendix~\ref{sec:appendiximple}.

\section{Conclusion}
We examine MoE from an ensemble perspective, highlighting how low-performing experts can adversely affect the overall performance of quantized MoE models.
Based on this insight, we propose Colla-Q, a bit-allocation framework based on an activation-entropy approach that maintains balanced performance across experts under quantization.
Unlike routing-based bit allocation strategies, Colla-Q estimates expert performance without token labels to reduce performance imbalance across experts.
We first establish an activation-entropy metric to estimate expert performance without token labels. Using this metric, we develop a minimax precision balancing algorithm that assigns more bits to low-performing experts.
In our experiments, Colla-Q minimizes quantization-induced performance loss while maintaining stable average performance across diverse MoE LLMs.
Moreover, our method shows smaller variation under calibration-domain shifts, indicating better generalization.
These results emphasize that, even in low-bit MoE quantization, maintaining balanced expert performance is important, as it supports collaborative expert contributions, thereby enhancing overall model performance and generalizability.

\section*{Limitations}
Our study demonstrates the effectiveness of the proposed method on several recent MoE language models under low-bit mixed-precision weight-only quantization. We evaluate Mixtral-8$\times$7B, DeepSeek-16B-Base, and Phi3.5-MoE at average bit-widths ranging from 1.57 to 2.54 bits. While the results show consistent performance across MoE architectures and quantization settings, the empirical scope remains limited in two aspects.

First, our experiments focus on weight-only quantization. Extending Colla-Q to joint weight–activation quantization would be a meaningful direction, as balancing expert performance is not inherently limited to weight quantization. Second, our experiments are limited to three representative MoE backbones—Mixtral-8$\times$7B, DeepSeek-MoE-16B-Base, and Phi3.5-MoE—with model sizes ranging from 16B to 46.7B parameters. These models were selected because they are widely used in prior MoE quantization and compression studies, enabling comparison under comparable settings. However, evaluation on newer MoE architectures and substantially larger models, such as those exceeding 100B parameters, remains limited by the considerable computational cost.

As an additional practical consideration, our expert-computation analysis shows improved efficiency with low-bit expert GEMM kernels, while the evaluation does not cover MoE-specific runtime overheads such as expert dispatch, memory movement, and small-GEMM scheduling. Since inference speed is as important as memory reduction in practice, end-to-end kernel and runtime optimization for mixed-precision MoE inference remains important. Taken together, extending Colla-Q to joint weight–activation quantization, broader and larger MoE architectures, and end-to-end optimization remain important directions for future work.

\section*{Acknowledgements}
This research was supported by the National Research Foundation of Korea (NRF), Electronics and Telecommunications Research Institute(ETRI), and Institute of Information \& Communications
Technology Planning \& Evaluation (IITP), funded by the Korean government [26CS1100, Development of Proprietary Physical AI-based Small-scale Computers and Integrated Soft Suits], and the Korea government(MSIT) (RS-2024-00356486, RS-2026-25617480, and IITP-2026-RS-2023-00255968).

\bibliography{citation}
\clearpage
\appendix
\section*{Appendix}
\setcounter{section}{1}

\subsection{Empirical Analysis of the Approximately Uniform Routing Weights}
\label{sec:a1_routingweights}
In Section~\ref{sec:moeensemble}, we assumed that the routing weights are approximately uniform in expectation over many tokens due to the load-balancing objective. To verify this assumption, we analyze the routing behavior of Mixtral 8$\times$7B on 128 samples from C4 with sequence length 2048. Since Mixtral 8$\times$7B adopts top-2 routing, $|\hat{E}(x)|=2$ for each token, and the reference value is $1/|\hat{E}(x)| = 0.5$. For each layer, we collect the router logits, identify the top-2 experts for each token, and compute the conditional average gate $\bar{g}_{e}$ of each expert as:
\begin{equation}
\bar{g}_{e}
=
\frac{\sum_x g_{e}(x)\,\mathbf{1}[e \in \hat{E}(x)]}
{\sum_x \mathbf{1}[e \in \hat{E}(x)]}.
\end{equation}
We then compute the mean absolute difference (MAD) between $\bar{g}_e$ and $1/|\hat{E}(x)|$ across experts. A smaller MAD indicates that routing weights are closer to the approximately uniform assumption.

\gatingweight
\gatingweightmad
\gatingweightmadall

As shown in Tables~\ref{tab:gatingweight} and~\ref{tab:gatingweightmad}, for Mixtral 8$\times$7B, the expert-wise conditional average routing weights are close to the reference value, and the layer-wise MAD is small in most layers.
We further verify this assumption using the same analysis on DeepSeek-MoE-16B-Base and Phi3.5-MoE in Table~\ref{tab:gatingweightmadall}.
These results support the approximately uniform routing-weight assumption across the evaluated MoE architectures with different expert counts and top-$k$ routing structures.

\subsection{Empirical Analysis of the Gaussian Approximation of Expert Activations}
\label{sec:a2_gaussian}
\gaussianquantifyall

In Section~\ref{sec:activationentropy}, Eq.~\ref{eq:diff_entropy} holds when the expert activations follow the Gaussian approximation.  To verify this condition,
we collect expert output activations from the evaluated MoE models using 128 samples from C4, each with a sequence length of 2048. We conduct two analyses corresponding to $H_{\text{within}}$ and $H_{\text{total}}$ in Eq.~\ref{eq:ratio_metric}: channel-wise Gaussianity of $A_{t,c}$ and expert-wise Gaussianity of pooled activations. For the channel-wise analysis, we sample 32 channels from each expert over four random seeds, while for the expert-wise analysis, we treat all activations from each expert as a single pooled distribution. We compare the distributions with the standard Gaussian using full-range $\ell_1$ distance and the fraction of units below a fixed $\ell_1$ threshold.

As shown in Table~\ref{tab:gaussianquantify}, the activation distributions remain close to the standard Gaussian across both channel-wise and expert-wise analyses. Figure~\ref{fig:gaussianplot} further shows that both the standardized channel-wise activations and the expert-wise pooled activations exhibit Gaussian-like bell-shaped distributions. This indicates that the dominant structure of expert activations is well captured by the Gaussian approximation. These results support the assumption used in Eq.~\ref{eq:diff_entropy}.
\gaussianplot

\subsection{Empirical Analysis of Activation-Entropy as an Expert Performance Proxy}
\label{sec:santiycheck}
To examine whether the proposed activation-entropy metric $\rho$ provides a meaningful proxy for expert performance, we conduct a controlled expert-pair intervention. The observed negative relationship between activation entropy and the expert performance estimated through this intervention supports its use as a proxy for expert performance. Since expert quality in MoE does not have an observable ground truth, we assess expert performance operationally through downstream performance under controlled expert combinations.

Specifically, we randomly select three layers of Mixtral 8$\times$7B, keep all other layers unchanged, and evaluate all 28 expert pairs in each target layer with routing weights fixed to 0.5 for equal contribution. We measure the downstream performance of each pair and define each expert's performance as the average over the seven pairs containing that expert. We then compute the Spearman correlation between $\rho$ and the resulting expert-level performance scores. As shown in Table~\ref{tab:expert_pair}, $\rho$ exhibits negative correlations across all three layers, with an average Spearman correlation of $-0.56$. Since lower $\rho$ indicates better estimated expert performance, this consistent negative correlation supports the intended ordering of our metric and its use as a label-free allocation proxy for relative expert performance.
\expertpair

\subsection{Additional Experimental Setup}
\label{sec:appendiximple}

\textbf{Baselines.}
Uniform quantization does not consider importance differences in the MoE structure, and instead assigns the same bit-width
to all experts.
BSP and OA-GPTQ~\citep{li2024quantmoe} are MoE-block-wise and expert-linear-layer-wise bit-width allocation methods, respectively. BSP estimates the importance of each MoE block using a separately trained predictor based on the cosine similarity of activations. OA-GPTQ computes linear-layer importance within each expert by scoring weight outliers.
MxMoE~\citep{duanmu2025mxmoe} estimates the importance of each linear layer within an expert using quantization sensitivity and expert activation patterns.
PMQ~\citep{huang2025mixture} is an expert-wise bit-width allocation method, which uses a routing-based metric for bit-width allocation.

\textbf{Implementation Details of Baselines.} BSP and OA-GPTQ use Wikitext2 for calibration with 128 sequences of length 2048. BSP allocates 4-bit to the top 25\% most important MoE blocks and 2-bit to the remaining blocks. OA-GPTQ allocates 4-bit to the top 25\% most important linear layers within experts and 2-bit to the rest. Both achieve a target average bit-width of approximately 2.54 bits (in DeepSeek, shared-expert layers receive 4-bit only if they fall into the selected top blocks/linear layers).
We reproduce the experimental results in model configurations that baseline methods~\citep{li2024quantmoe} do not report.

MxMoE uses Wikitext2 for calibration with 128 sequences, and we set the sequence length to 2048. We include all linear layers in the bit-width allocation, including those in the shared expert of DeepSeek, and we do not manually fix the shared-expert bit-width. For a fair comparison, we quantize the attention and gating modules to 4-bit, consistent with our baselines. In low-bit settings, following the MxMoE setup for reproduction, we set the runtime-cost weighting parameter $r$ to 1. Therefore, we can perform a fair comparison between MxMoE and ours under the same conditions.

\TabModelInfo
\mmlufiveshotfull

\deepseektotaltable
\phitotaltable
\calibrationheatmap
\calibrationcosine

\calibdatasetablationcombined

PMQ and our method use C4 as the calibration dataset with 128 sequences. For each MoE block, we apply GPTQ-based 4-bit quantization to the attention and gating modules. For DeepSeek-16B-Base, since the first layer is a dense MLP rather than an MoE block, we quantize it using 4-bit per-channel GPTQ as well. For models with shared experts, we fix the shared-expert bit-width to 2-bit when targeting average bit-widths of 1.57 and 2.05 bits, and to 4-bit for the 2.54-bit setting. Model architectures (e.g., the number of layers, experts, and Top-$k$ routing) are summarized in Table~\ref{tab:model-info}, and we follow these configurations when applying the above quantization settings.

\textbf{Expert-Computing Efficiency Evaluation.} Following the kernel-level efficiency evaluation of MxMoE~\citep{duanmu2025mxmoe}, we measure expert computation. We use the actual routing results of Mixtral 8$\times$7B on 128 WikiText2 sequences with a sequence length of 512 and execute each expert according to its routed token count. The reported average bit-widths (e.g., 2.54 bits) denote effective model-level bit-widths obtained from discrete expert-wise bit assignments, with the attention and gating modules fixed at 4 bits, rather than fractional-bit kernels.
Based on the bit-width assigned by Colla-Q, experts are executed using INT1$\times$FP16, INT2$\times$FP16, or INT3$\times$FP16 weight-only GEMM kernels and compared with FP16 under the same routing workload. Measurements are conducted on a single NVIDIA RTX 3090 24GB GPU with 10 warm-up iterations and 30 repetitions. Attention and MoE routing/dispatch are excluded, and TFLOPS is computed from the expert-MLP FLOPs and measured kernel execution time.

\subsection{Additional Experimental Results}
\label{sec:fullresults}
\noindent \textbf{Complete MMLU Few-shot Results.}
Table~\ref{tab:mmlufiveshotfull} extends Table~\ref{tab:mmlufiveshot}  by reporting MMLU 5-shot performance for Mixtral 8$\times$7B at additional average bit-widths (2.05-bit and 1.57-bit), together with the 2.54-bit setting.

\noindent \textbf{Complete Results Across All Benchmarks.}
We provide complete experimental results to complement Table~\ref{tab:deepsk-phi} of the manuscript.
Table~\ref{tab:deepseek_total} and Table~\ref{tab:phi_total} show the complete experimental results on eight benchmarks for DeepSeek-16B-Base and Phi3.5-MoE across different average bit-widths and MoE quantization methods.

\noindent \textbf{Calibration Dataset Dependence.}
We provide experimental results regarding the calibration dataset dependence.
Figure~\ref{fig:calibrationheatmap} visualizes the heatmaps of the routing-based and activation-entropy metrics across experts and MoE blocks for the two models Mixtral 8$\times$7B and Phi3.5-MoE.
They demonstrate that the activation-entropy metric produces similar patterns of the heatmaps across different calibration datasets.
This result implies that our method operates without dependence on the particular calibration dataset, yielding generalized quantization results.
Figure~\ref{fig:calibrationcosine} provides the complete cosine-similarity heatmaps corresponding to the summary statistics reported in Table~\ref{tab:calb_sim_avg}.
In our experiments, we compute the cosine similarity as:

\begin{equation}
\label{eq:mean_layer_cos_sim}
S=
\frac{1}{|\mathcal{L}|}
\sum_{l\in\mathcal{L}}
\frac{\mathbf{a}_l^\top \mathbf{b}_l}{\|\mathbf{a}_l\|_2\,\|\mathbf{b}_l\|_2},
\end{equation}
where $\mathcal{L}$ denotes the set of MoE blocks, $\mathbf{a}_l$ and $\mathbf{b}_l$ denote the metric vectors (\eg activation-entropy metric).
Using this cosine similarity across MoE blocks, we can confirm that our activation-entropy metric produces consistent results across different calibration datasets, as shown in Figure~\ref{fig:calibrationcosine}. Furthermore, Table~\ref{tab:calb_combined} reports the corresponding zero-shot performance
under each calibration dataset for both models.
\algorithmmetirctable
\dilpargo
\quantlosscomparefull
\linearexpertcomparefull

\subsection{Additional Ablation Studies} \label{sec:additional_ablation}
\textbf{Ablation of Metric and Allocation Strategy.} To isolate the contributions of the metric and allocation strategy, we conduct a $2\times2$ ablation by cross-applying the PMQ routing-based and our activation-entropy metrics with the PMQ ILP allocation and our minimax precision balancing. As shown in Table~\ref{tab:algorithmmetirctable}, minimax precision balancing consistently outperforms ILP with the activation-entropy metric across all bit-widths, while the gains are less consistent with the PMQ metric. These results support the complementary roles of activation entropy and minimax precision balancing in Colla-Q.

\textbf{Ablation of the Bit-Allocation Procedure.} To evaluate the effectiveness of our minimax precision balancing procedure, we compare it with Direct ILP. For a fair comparison, we reformulate Direct ILP to optimize the same Colla-Q minimax objective, rather than the conventional sum-based objective, under the same bit budget. While Direct ILP determines the bit-width configuration through a single optimization, our method recursively allocates additional bits to the worst-performing expert. As shown in Table~\ref{tab:directilp}, minimax precision balancing consistently achieves higher downstream performance across all bit-widths. These results support the effectiveness of our precision-balancing procedure for discrete expert-wise bit allocation.

\subsection{Complete Results for Ablation Study}
\label{sec:complete_ablation}
We offer the complete experimental results of our ablation studies in Tables~\ref{tab:quantlosstable} and~\ref{tab:linearexpert} of the manuscript. Tables~\ref{tab:quantlossfull} and~\ref{tab:linearexpertfull} further extend these results by reporting the full benchmark outcomes for activation-entropy-weighted quantization error and expert-wise mixed-precision allocation. These additional results allow a more detailed assessment of the consistency of the observed trends across tasks.

\end{document}

%% file: figure_table.tex
\usepackage{amsmath}
\usepackage{threeparttable}
\usepackage{array}

\newcommand{\introfigureTwo}{
\begin{figure}[t]
\centering
\subfloat[\centering Accuracy of ours and baseline]{\includegraphics[width=0.999\linewidth]{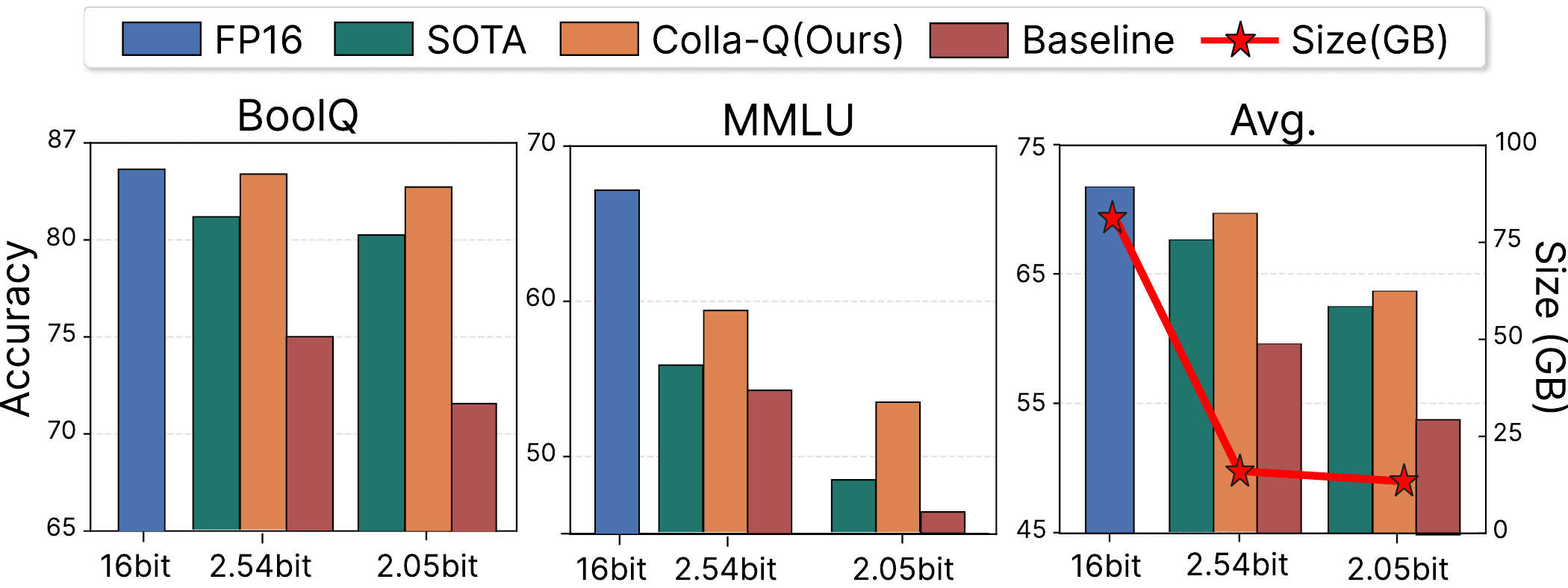}\label{fig:compare_baseline}}
\vfill
\subfloat[\centering Performance across experts after quantization]{\includegraphics[width=0.999\linewidth]{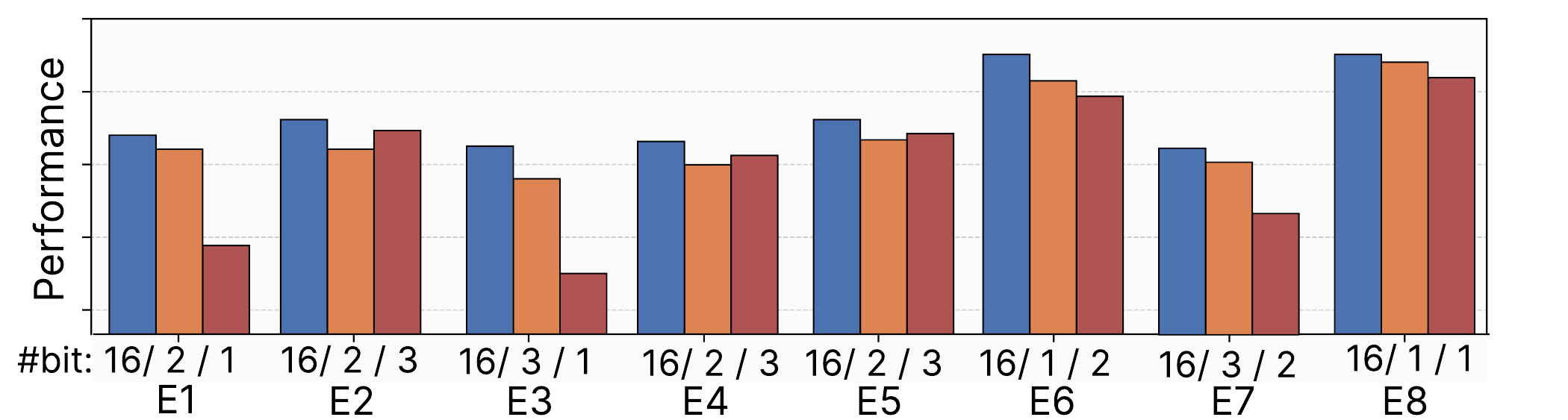}\label{fig:compare_balance}}\caption {
Preliminary study supporting our approach.
The baseline refers to the quantized model with randomly assigned expert bit-widths.
(a) Zero-shot performance (\%) of Mixtral 8$\times$7B under our method and baseline across different bit-widths.
(b) The baseline results in imbalanced expert performance after quantization, whereas our method allocates more bits to preserve similar performance across experts. Strong experts tend to degrade less at low bit-widths. The x-axis shows the experts and their assigned bit-widths, and the y-axis shows the expert performance measured by our metric.
}
\label{fig:introfiguretwo}
\end{figure}
}

\newcommand{\introfigure}{
\begin{figure*}[t]
\centering
\subfloat[\centering Baseline]{\includegraphics[width=0.49\linewidth]{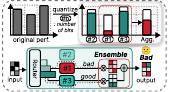}\label{fig:moe-arch}}
\hfill
\subfloat[\centering Our approach]{\includegraphics[width=0.49\linewidth]{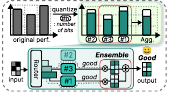}\label{fig:intro-concpet}}
\caption {
Conceptual illustration of the proposed method. (a) Allocating fewer bits to a lower-performing expert can significantly degrade its performance, leading to suboptimal performance of the aggregated output of the MoE block. (b) In contrast, our method allocates more bits to a lower-performing expert, thereby improving the aggregated output and preserving the overall quality of the MoE block output. Agg. denotes the aggregated MoE block output obtained by combining the outputs of routed experts.
}
\label{fig:introfigure}
\end{figure*}
}

\newcommand{\Mixtralcalb}{
\begin{figure*}[t]
    \centering
  \includegraphics[width=1.0\linewidth]{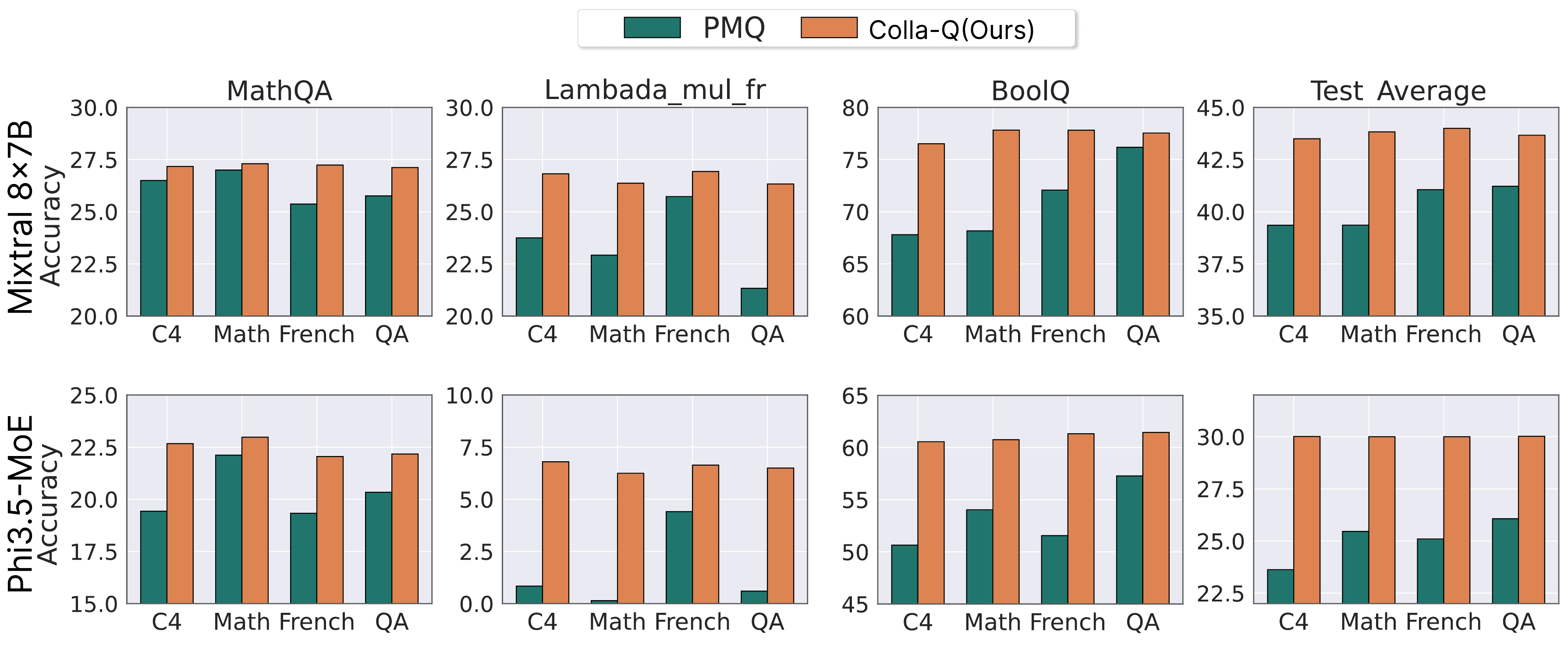}
    \caption{
    Performance comparison of PMQ and Ours on Mixtral 8$\times$7B and Phi3.5-MoE quantized to 1.57-bit with different calibration datasets (C4, Math, French, QA).
    Each subplot corresponds to a test dataset, the $x$-axis denotes the calibration dataset, and the $y$-axis shows zero-shot accuracy. Test Average denotes the mean accuracy over all three test datasets. }
  \label{fig:mixtralcalb}
\end{figure*}
}

\newcommand{\ablationcompute}{
\begin{figure}[t]
\centering
\includegraphics[width=0.75\linewidth]{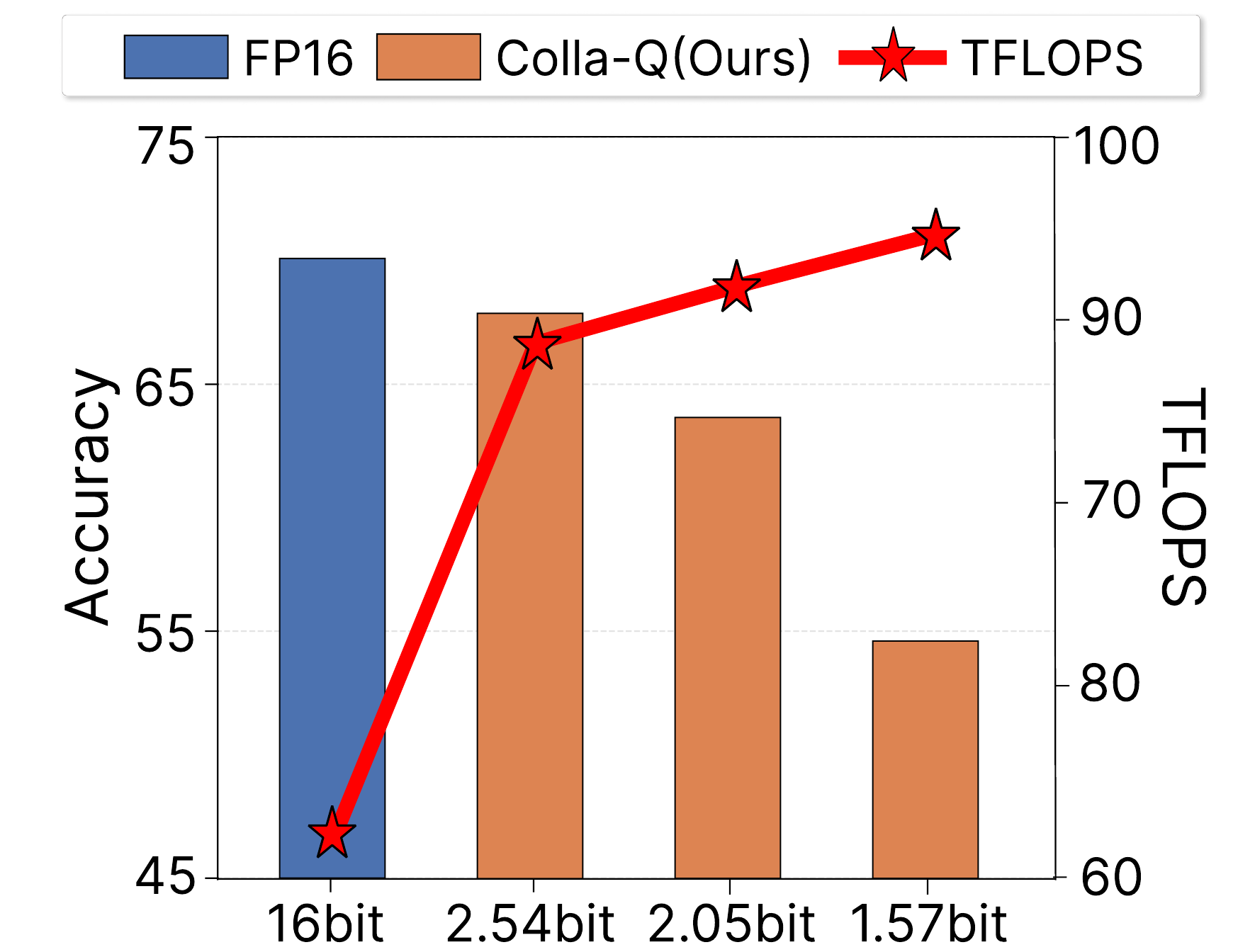}
\caption {
Expert-computing efficiency of Colla-Q on Mixtral 8$\times$7B. Bars show zero-shot accuracy across eight benchmarks, and the line shows TFLOPS under the routing workload. Average bit-widths denote allocation budgets with discrete expert-wise bit assignments.
}
\label{fig:ablationcompute}
\end{figure}
}

\newcommand{\oursalgorithmtwo}{
\begin{algorithm}[!bth]
\caption{Minimax precision balancing}
\label{alg:ours}
\begin{algorithmic}[1]

\State \textbf{Input} \\
    Expert set: $E = \{e_1, \dots, e_N\}$ \\
    Total budget and max bit-width: $B_{\text{total}}$, $b_{\max}$ \\
    Activation-entropy weights: $\{\rho_{e_1}, \dots, \rho_{e_N}\}$

\State \textbf{Output} \\
    Bit-width allocation $\mathbf{b}^* = \{b_{e_1}, \dots, b_{e_N}\}$

\Statex

\State \textbf{Initialization:} \\
    $\forall i,\; b_{e_i} \leftarrow 1$ \\
    $B_{\text{current}} \leftarrow \sum_{i=1}^{N} b_{e_i}$ \\

\While{$B_{\text{current}} < B_{\text{total}}$}
    \State $\mathcal{C} \leftarrow \{ i \mid b_{e_i} < b_{\max} \}$
    \Comment{Identify experts below max bit-budget}

    \If{$\mathcal{C} = \emptyset$}
        \State \textbf{break}
        \Comment{All experts reached $b_{\max}$}
    \EndIf

    \ForAll{$c \in \mathcal{C}$}
        \State $\mathcal{E}_c \leftarrow \|\,e_c(\mathbf{x})-\tilde{e}_c(\mathbf{x}; b_{e_c})\,\|_2^2$
        \State $\mathcal{L}_{c} \leftarrow \rho_{e_c} \times \mathcal{E}_c$
    \EndFor

    \State $k^* \leftarrow \operatorname*{argmax}_{c \in \mathcal{C}} (\mathcal{L}_{c})$ \Comment{Find the worst performing expert}

    \State $b_{e_{k^*}} \leftarrow b_{e_{k^*}} + 1$ \Comment{Allocate 1 bit}
    \State $B_{current} \leftarrow B_{current} + 1$
\EndWhile

\State \Return $\mathbf{b}^*$
\end{algorithmic}
\end{algorithm}
}

\newcommand{\gaussianplot}{
\begin{figure}[t]
\centering
\includegraphics[width=1.0\linewidth]{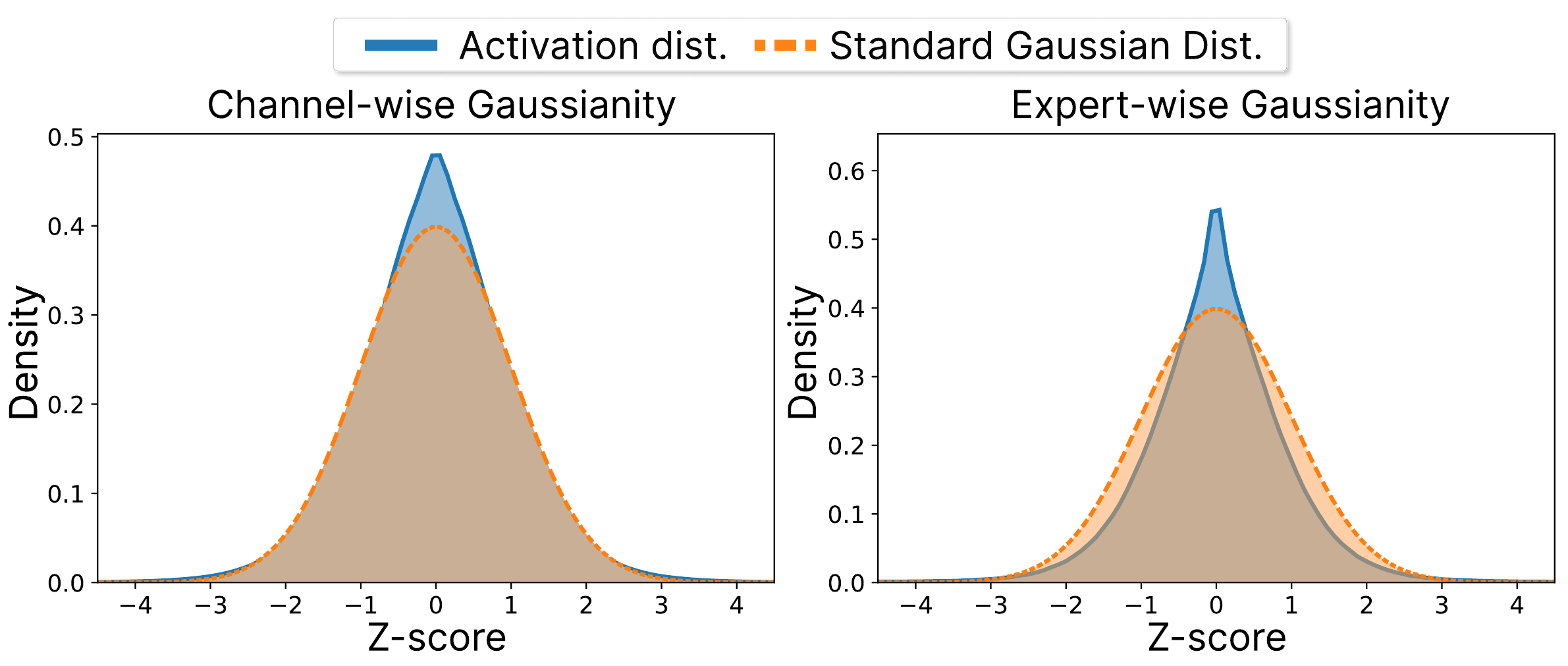}
\caption {Channel-wise and expert-wise Gaussianity of expert FFN output activations in Mixtral 8$\times$7B. Activation dist. denotes the averaged standardized density over sampled channels (left) and the standardized density of pooled expert activations (right).}
\label{fig:gaussianplot}
\end{figure}
}

\newcommand{\Othersmodeltalbe}{
\definecolor{charcoal}{RGB}{54,69,79}
\colorlet{oursrow}{charcoal!12}
\begin{table*}[t]
\centering
\small
\setlength{\tabcolsep}{15pt}
\renewcommand{\arraystretch}{0.97}
\definecolor{charcoal}{RGB}{54,69,79}
\begin{tabular}{c c c c c c c c }
\toprule
\midrule
\textbf{\multirow{2}{*}{Bits}} & \textbf{\multirow{2}{*}{Method}}
 &\multicolumn{3}{c}{\textbf{DeepSeek-16B-Base}} & \multicolumn{3}{c}{\textbf{Phi3.5-MoE}} \\
\cmidrule(lr){3-5}\cmidrule(lr){6-8}
& & \textbf{MMLU} & \textbf{C.S Avg.} & \textbf{Avg. }  &
\textbf{MMLU} & \textbf{C.S Avg.} & \textbf{Avg. } \\
\midrule
16 & - & 38.0 & 65.1 & 51.6 & 76.6 & 68.2 & 72.4  \\
\midrule
\multirow{5}{*}{2.54}
& OA-GPTQ & 33.3 & 61.9 & 47.6 & 43.8 & 53.4 & 48.6 \\
& BSP & 29.7 & 56.7 & 43.2& 39.4 & 44.9 & 42.2  \\
 & MxMoE & 29.8 & 62.1 & 46.0 & 43.9 & 51.3 & 47.6\\
 & PMQ & 32.8 & 61.9 & 47.3  & 50.7 & 57.8 & 54.2  \\
 & {\cellcolor{charcoal!12}}Colla-Q &
 {\cellcolor{charcoal!12}}33.4 &
 {\cellcolor{charcoal!12}}63.2 &
 {\cellcolor{charcoal!12}}48.3 &
 {\cellcolor{charcoal!12}}51.8 &
 {\cellcolor{charcoal!12}}58.4 &
 {\cellcolor{charcoal!12}}55.1  \\
\midrule
\multirow{3}{*}{2.05}
 & MxMoE & 26.4 & 57.8 & 42.1  & 27.8 & 42.3 & 35.1\\
& PMQ & 27.0 & 58.0 & 42.5  & 25.4 & 50.5 & 38.0  \\
 & {\cellcolor{charcoal!12}}Colla-Q &
 {\cellcolor{charcoal!12}}27.1 &
 {\cellcolor{charcoal!12}}59.2 &
 {\cellcolor{charcoal!12}}43.1 &
 {\cellcolor{charcoal!12}}28.3 &
 {\cellcolor{charcoal!12}}52.9 &
 {\cellcolor{charcoal!12}}40.6  \\
\midrule
\multirow{3}{*}{1.57}
 & MxMoE & 24.0 & 45.4 & 34.7 & 23.7  &  38.8 & 31.3 \\
& PMQ & 22.6 & 54.0 & 38.3 & 23.5 & 41.8 & 32.7 \\
 & {\cellcolor{charcoal!12}}Colla-Q &
 {\cellcolor{charcoal!12}}23.3 &
 {\cellcolor{charcoal!12}}54.7 &
 {\cellcolor{charcoal!12}}39.0 &
 {\cellcolor{charcoal!12}}24.3 &
 {\cellcolor{charcoal!12}}43.3 &
 {\cellcolor{charcoal!12}}33.8 \\
\midrule
\bottomrule
\end{tabular}
\caption{Zero-shot performance comparison of two MoE models (DeepSeek-16B-Base and Phi3.5-MoE) under different average bit-widths and MoE quantization methods. C.S Avg. denotes the mean accuracy over the seven non-MMLU benchmarks. Avg. is the averaged value of MMLU and C.S Avg. Full results can be found in Appendix~\ref{sec:fullresults}. }
\label{tab:deepsk-phi}
\end{table*}
}

\newcommand{\Mixtraltable}{
\definecolor{charcoal}{RGB}{54,69,79}
\begin{table*}[t]
\centering
\small
\setlength{\tabcolsep}{7pt}
\renewcommand{\arraystretch}{0.97}
\begin{tabular}{ccccccccccccc}
\toprule
\midrule
\textbf{Bits} & \textbf{Method} &
\textbf{MMLU} & \textbf{PIQA} & \textbf{ARC-e} & \textbf{ARC-c} &
\textbf{BoolQ} & \textbf{HellaS.} & \textbf{Wino.} & \textbf{MathQA} &
\textbf{Avg. ($\uparrow$)}  \\
\midrule
16  & - & 67.8 & 83.6 & 84.2 & 56.5 & 85.0 & 84.0 & 76.2 & 41.7 & 72.4 \\
\midrule
2 & Uniform & 30.2 & 60.7 & 47.1 & 25.7 & 62.3 & 41.9 & 52.8 & 22.4 & 42.9  \\
\midrule
\multirow{5}{*}{2.54} &  OA-GPTQ & 30.3 & 69.6 & 57.0 & 34.7 & 58.1 & 59.9 & 61.3 & 26.6 & 49.7  \\
   & BSP & 25.1 & 62.8 & 49.4 & 29.8 & 53.8 & 51.5 & 56.0 & 23.9 & 44.0  \\
   & MxMoE & 54.4 & 79.7 & 74.8 & 50.0 & 79.1 & 78.1 & 71.2 & 33.1 & 65.1  \\
   & PMQ & 56.4 & 80.5 & 77.1 & 51.3 & 82.5 & 79.0 & 74.0 & 39.2 & 67.5  \\
   & {\cellcolor{charcoal!12}}Colla-Q & {\cellcolor{charcoal!12}}59.4 & {\cellcolor{charcoal!12}}80.5 & {\cellcolor{charcoal!12}}79.8 & {\cellcolor{charcoal!12}}54.0 & {\cellcolor{charcoal!12}}84.7 & {\cellcolor{charcoal!12}}78.5 & {\cellcolor{charcoal!12}}74.9 & {\cellcolor{charcoal!12}}36.5 & {\cellcolor{charcoal!12}}68.5  \\
\midrule
\multirow{3}{*}{2.05} & MxMoE & 42.3 & 75.4 & 68.6 & 41.5 & 71.4 & 71.7 & 66.8 & 29.9 & 58.4 \\
   & PMQ & 46.8 & 79.2 & 73.1 & 48.4 & 80.6 & 75.0 & 71.3 & 31.8 & 63.3  \\
   & {\cellcolor{charcoal!12}}Colla-Q & {\cellcolor{charcoal!12}}52.9 & {\cellcolor{charcoal!12}}77.8 & {\cellcolor{charcoal!12}}74.9 & {\cellcolor{charcoal!12}}46.4 & {\cellcolor{charcoal!12}}83.3 & {\cellcolor{charcoal!12}}72.4 & {\cellcolor{charcoal!12}}72.0 & {\cellcolor{charcoal!12}}32.3 & {\cellcolor{charcoal!12}}64.0  \\
\midrule
\multirow{3}{*}{1.57} &  MxMoE & 28.5 & 61.7 & 49.5 & 29.8 & 64.1 & 46.2 & 60.0 & 24.5 & 45.5 \\
   & PMQ & 32.3 & 72.4 & 62.5 & 37.9 & 73.6 & 63.2 & 66.4 & 26.8 & 54.5 \\
   & {\cellcolor{charcoal!12}}Colla-Q & {\cellcolor{charcoal!12}}36.6 & {\cellcolor{charcoal!12}}71.0 & {\cellcolor{charcoal!12}}63.1 & {\cellcolor{charcoal!12}}36.2 & {\cellcolor{charcoal!12}}75.8 & {\cellcolor{charcoal!12}}59.8 & {\cellcolor{charcoal!12}}66.2 & {\cellcolor{charcoal!12}}26.9 & {\cellcolor{charcoal!12}}54.5  \\
\midrule
\bottomrule
\end{tabular}
\caption{Zero-shot task performance (\%) of the quantized Mixtral 8$\times$7B on eight benchmarks under different average bit-widths and MoE quantization methods. Bits denotes the average effective bit-widths for each quantization method. Avg. reports the averaged accuracy over all tasks.}
\label{tab:mixtral87b_main}
\end{table*}}

\newcommand{\quantlosspmqours}{
\begin{table*}
\centering
\small
\setlength{\tabcolsep}{8.3pt}
\renewcommand{\arraystretch}{1.2}
\definecolor{charcoal}{RGB}{54,69,79}
\begin{tabular}{ccccccccccc}
\toprule
\midrule
\textbf{Bits} & \textbf{Metric} & \textbf{MMLU} & \textbf{PIQA} & \textbf{ARC-e} & \textbf{ARC-c} & \textbf{BoolQ} & \textbf{HellaS.} & \textbf{Wino.} & \textbf{MathQA} & \textbf{Avg.} \\
\midrule
\multirow{2}{*}{2.54} & PMQ & 58.6 & 81.2 & 78.0 & 52.2 & 82.2 & 79.5 & 74.0 & 35.7 & 67.7  \\
 & {\cellcolor{charcoal!12}}Colla-Q & {\cellcolor{charcoal!12}}59.4 & {\cellcolor{charcoal!12}}80.5 & {\cellcolor{charcoal!12}}79.8 & {\cellcolor{charcoal!12}}54.0 & {\cellcolor{charcoal!12}}84.7 & {\cellcolor{charcoal!12}}78.5 & {\cellcolor{charcoal!12}}74.9 & {\cellcolor{charcoal!12}}36.5 & {\cellcolor{charcoal!12}}68.5 \\
 \midrule
\multirow{2}{*}{2.05} & PMQ & 45.5 & 77.6 & 74.9 & 47.2 & 82.8 & 74.8 & 71.6 & 32.0 & 63.3  \\
 &{\cellcolor{charcoal!12}}Colla-Q & {\cellcolor{charcoal!12}}52.9 & {\cellcolor{charcoal!12}}77.8 & {\cellcolor{charcoal!12}}74.9 & {\cellcolor{charcoal!12}}46.4 & {\cellcolor{charcoal!12}}83.3 & {\cellcolor{charcoal!12}}72.4 & {\cellcolor{charcoal!12}}72.0 & {\cellcolor{charcoal!12}}32.3 & {\cellcolor{charcoal!12}}64.0  \\
 \midrule
\multirow{2}{*}{1.57} & PMQ & 33.1 & 70.3 & 63.9 & 36.8 & 71.4 & 59.8 & 67.2 & 26.5 & 53.6  \\
 &{\cellcolor{charcoal!12}}Colla-Q & {\cellcolor{charcoal!12}}36.6 & {\cellcolor{charcoal!12}}71.0 & {\cellcolor{charcoal!12}}63.1 & {\cellcolor{charcoal!12}}36.2 & {\cellcolor{charcoal!12}}75.8 & {\cellcolor{charcoal!12}}59.8 & {\cellcolor{charcoal!12}}66.2 & {\cellcolor{charcoal!12}}26.9 & {\cellcolor{charcoal!12}}54.5\\
 \midrule
 \bottomrule
\end{tabular}
\caption{Zero-shot performance of Mixtral 8$\times$7B on eight benchmarks under different average bit-widths. In comparison, we use PMQ’s routing-based metric and activation-entropy metric with minimax precision balancing.}
\label{tab:quantlosspmqours}
\end{table*}}

\newcommand{\calbfactorcos}{
\begin{table}
\centering
\small
\setlength{\tabcolsep}{10.pt}
\renewcommand{\arraystretch}{1.1}
\begin{tabular}{ccccc}
\toprule
\midrule
 \multirow{2}{*}{\textbf{\begin{tabular}[c]{@{}c@{}}Calibration \\dataset\end{tabular}}}& \multicolumn{2}{c}{\textbf{Mixtral 8$\times$7B}} & \multicolumn{2}{c}{\textbf{Phi3.5-MoE}} \\
\cmidrule(lr){2-3}\cmidrule(lr){4-5}
 &  PMQ & Ours & PMQ & Ours \\
\midrule
C4 & 86.0 & 98.8 & 72.1 & 94.0 \\
Math & 85.5 & 98.5 & 46.3 & 91.7 \\
French & 86.0 & 98.2 & 57.8 & 91.6 \\
QA & 72.0 & 98.9 & 72.7 & 94.8 \\
\midrule
\bottomrule
\end{tabular}
\caption{Average cosine similarity of routing-based metric from PMQ and our metric. For each calibration dataset (row), we compute the mean cosine similarity to the other three calibration datasets, excluding self-similarity. Details are provided in Appendix~\ref{sec:fullresults}.}
\label{tab:calb_sim_avg}
\end{table}}

\newcommand{\linearexpertcomparison}{
\begin{table}
\centering
\small
\setlength{\tabcolsep}{7.5pt}
\renewcommand{\arraystretch}{1.05}
\definecolor{charcoal}{RGB}{54,69,79}
\begin{tabular}{ccccc}
\toprule
\midrule
\textbf{Bits} & \textbf{Components} & \textbf{MMLU} & \textbf{C.S Avg.} & \textbf{Avg.}  \\
\midrule
16 &  - & 67.8 & 73 & 70.4 \\
\midrule
\multirow{2}{*}{2.54} & Linear Layer & 57.9 & 69.9 & 63.9  \\
 & {\cellcolor{charcoal!12}}Expert &  {\cellcolor{charcoal!12}}59.4 &
 {\cellcolor{charcoal!12}}69.8 &
 {\cellcolor{charcoal!12}}64.6 \\
 \midrule
\multirow{2}{*}{2.05} & Linear Layer & 49.1 & 65.1 & 57.1  \\
 & {\cellcolor{charcoal!12}}Expert &  {\cellcolor{charcoal!12}}52.9 &
 {\cellcolor{charcoal!12}}65.6 &
 {\cellcolor{charcoal!12}}59.3  \\
 \midrule
\multirow{2}{*}{1.57} & Linear Layer & 34.1 & 54.3 & 44.2 \\
 & {\cellcolor{charcoal!12}}Expert &  {\cellcolor{charcoal!12}}36.6 &
 {\cellcolor{charcoal!12}}57.0 &
 {\cellcolor{charcoal!12}}46.8  \\
\midrule
\bottomrule
\end{tabular}
\caption{Ablation study comparing our minimax precision balancing when applied to different components of Mixtral 8$\times$7B. We compare expert-level allocation with linear-layer-level allocation across different average bit-width settings under identical conditions.}
\label{tab:linearexpert}
\end{table}}

\newcommand{\mmlufiveshot}{
\begin{table}[t]
\centering
\small
\setlength{\tabcolsep}{22.pt}
\renewcommand{\arraystretch}{1}
\definecolor{charcoal}{RGB}{54,69,79}
\begin{tabular}{ccc}
\toprule
\midrule
\textbf{Bits} & \textbf{Method} & \textbf{MMLU}  \\
\midrule
16 & - & 70.6 \\
\midrule
\multirow{5}{*}{2.54} & OA-GPTQ & 58.1  \\
 & BSP & 51.7  \\
 & MxMoE & 59.1  \\
 & PMQ & 61.2 \\
 & {\cellcolor{charcoal!12}}Colla-Q & {\cellcolor{charcoal!12}}62.5 \\
 \midrule
 \bottomrule
\end{tabular}
\caption{MMLU 5-shot task performance of Mixtral 8$\times$7B under different MoE quantization methods at an average bit-width of 2.54. Additional results under other bit-width settings are provided in the Appendix~\ref{sec:fullresults}. }
\label{tab:mmlufiveshot}
\end{table}
}

\newcommand{\mmlufiveshotfull}{
\begin{table}[b]
\centering
\small
\setlength{\tabcolsep}{22.3pt}
\renewcommand{\arraystretch}{1}
\definecolor{charcoal}{RGB}{54,69,79}
\begin{tabular}{ccc}
\toprule
\midrule
\textbf{Bits} & \textbf{Method} & \textbf{MMLU}  \\
\midrule
16 & - & 70.6 \\
\midrule
\multirow{5}{*}{2.54} & OA-GPTQ & 58.1  \\
 & BSP & 51.7 \\
 & MxMoE & 59.1 \\
 & PMQ & 61.2  \\
 & {\cellcolor{charcoal!12}}Colla-Q & {\cellcolor{charcoal!12}}62.5  \\
 \midrule
\multirow{3}{*}{2.05} & MxMoE & 50.2  \\
 & PMQ & 49.8 \\
 & {\cellcolor{charcoal!12}}Colla-Q & {\cellcolor{charcoal!12}}53.2   \\
 \midrule
\multirow{3}{*}{1.57} & MxMoE & 29.1  \\
 & PMQ & 33.4 \\
 & {\cellcolor{charcoal!12}}Colla-Q & {\cellcolor{charcoal!12}}34.1  \\
 \midrule
 \bottomrule
\end{tabular}
\caption{MMLU 5-shot task performance of Mixtral 8$\times$7B, comparing MoE quantization methods.}
\label{tab:mmlufiveshotfull}
\end{table}
}

\newcommand{\calibrationcosine}{
\begin{figure*}[t]
\centering
\subfloat[\centering Mixtral 8$\times$7B's factor cosine heatmap]{\includegraphics[width=0.46\linewidth]{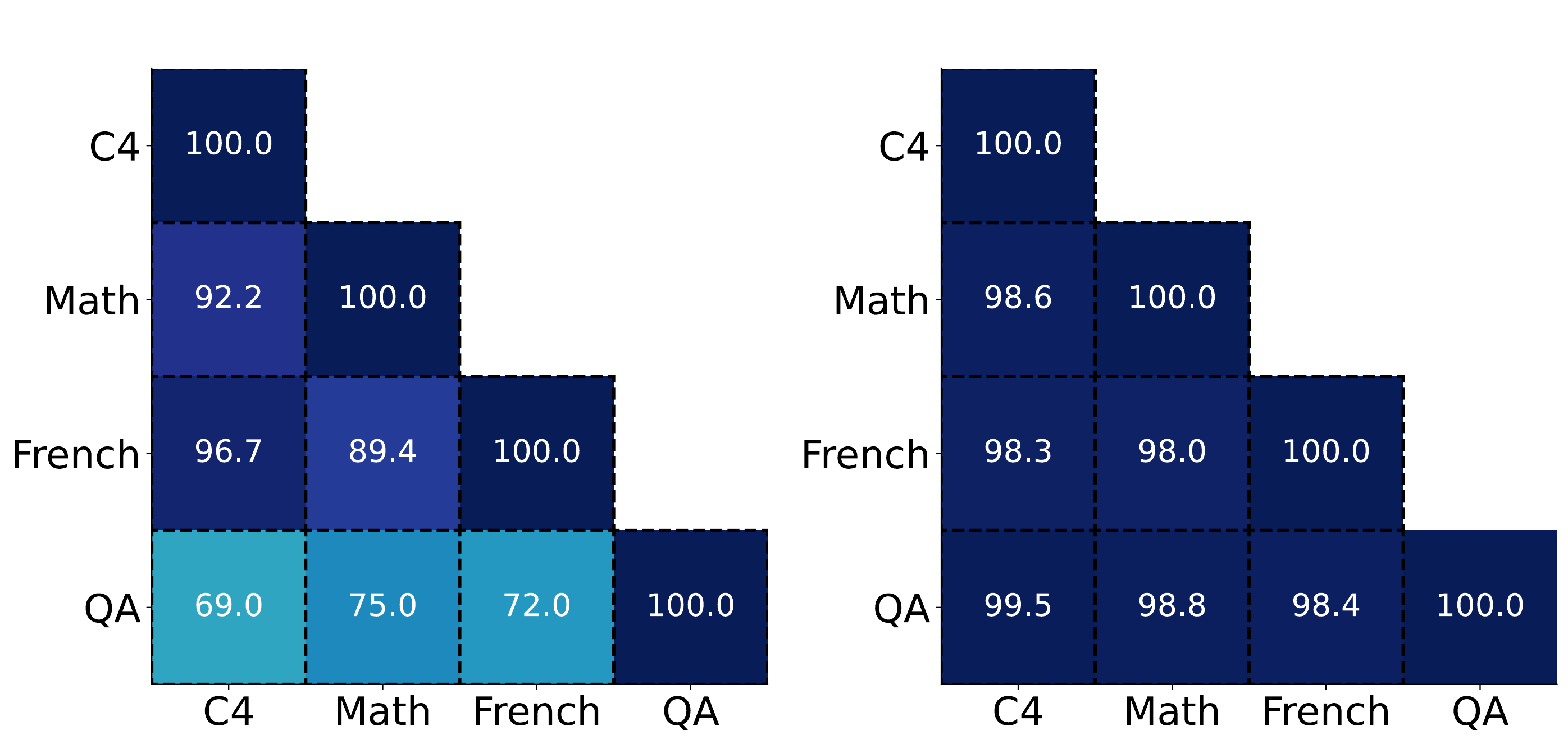}}
\subfloat[\centering Phi3.5-MoE's factor cosine heatmap]
{\includegraphics[width=0.54\linewidth]{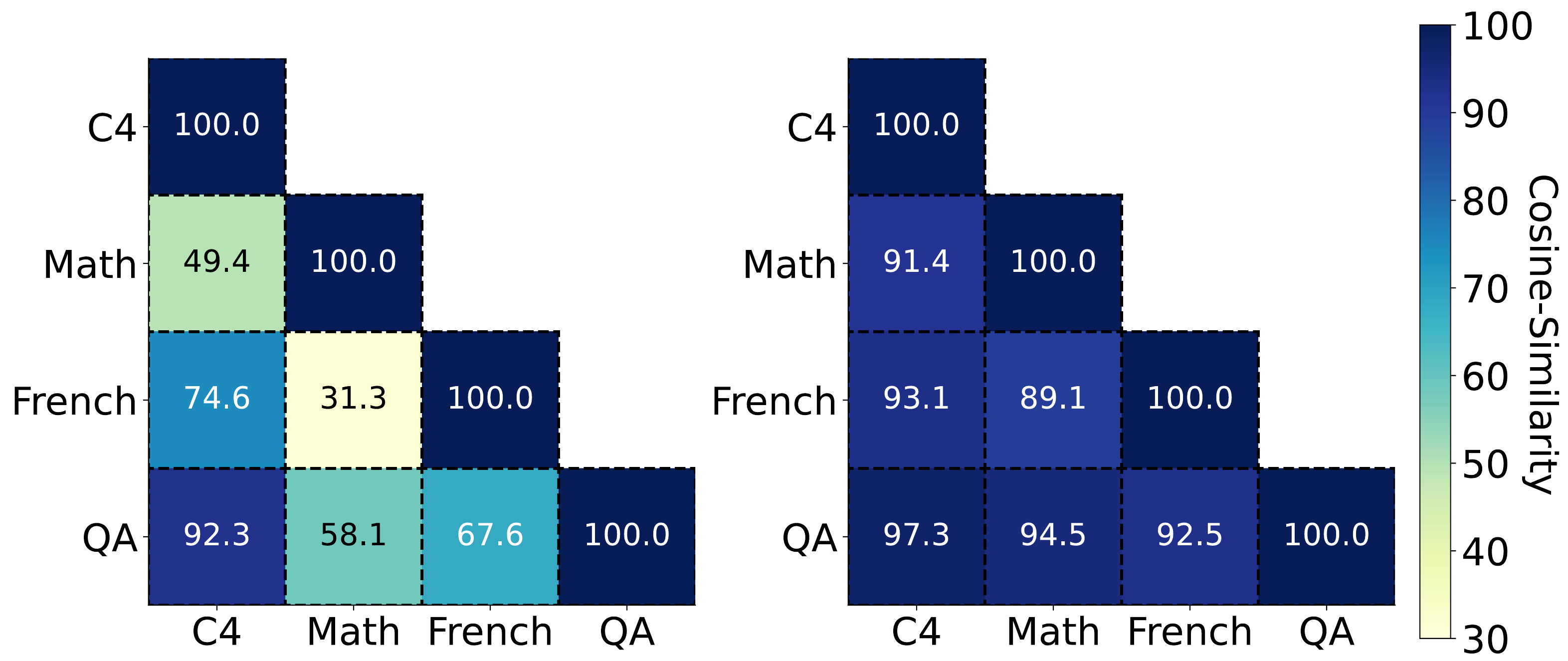}}
\caption{Cosine similarity heatmaps of routing-based and our metrics across calibration datasets on Mixtral 8×7B and Phi3.5-MoE models. For each model, the left heatmap shows similarities between metrics obtained with PMQ and the right reports those obtained with ours. Cosine similarities in [0, 1] are multiplied by 100.}
\label{fig:calibrationcosine}
\end{figure*}
}

\newcommand{\calibrationheatmap}{
\begin{figure*}[!bth]
\centering
\subfloat[\centering Mixtral 8$\times$7B's factor vectors]{\includegraphics[width=0.49\linewidth]{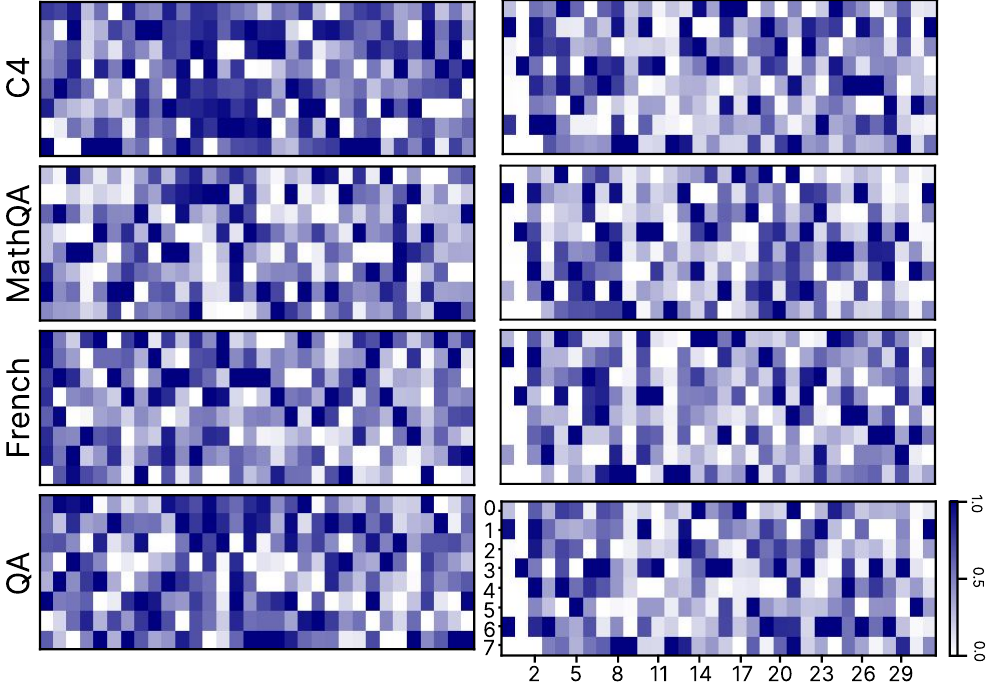}}
\hfill
\subfloat[\centering  Phi3.5-MoE's factor vectors]
{\includegraphics[width=0.49\linewidth]{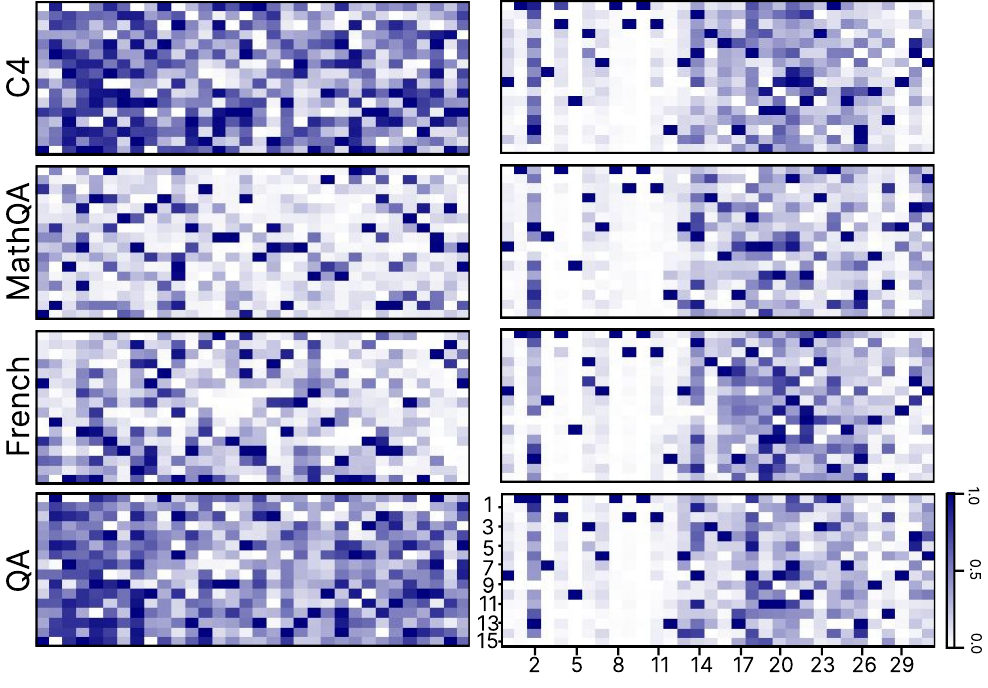}}
\caption{Heatmaps of the PMQ's routing-based metric (left) and our metric (right) computed from different calibration datasets on Mixtral 8×7B and Phi3.5-MoE models. For each calibration dataset, each heatmap shows the normalized values across MoE blocks.}
\label{fig:calibrationheatmap}
\end{figure*}
}

\newcommand{\deepseektotaltable}{
\begin{table*}[t]
\centering
\small
\setlength{\tabcolsep}{7.7pt}
\renewcommand{\arraystretch}{1}
\definecolor{charcoal}{RGB}{54,69,79}
\begin{tabular}{cccccccccccc}
\toprule
\midrule
 \textbf{Bits}  & \textbf{Method} & \textbf{MMLU} & \textbf{PIQA} & \textbf{ARC-e} & \textbf{ARC-c} & \textbf{BoolQ} & \textbf{HellaS.} & \textbf{Wino.} & \textbf{MathQA} & \textbf{Avg.}   \\
\midrule
 16 & - & 38.0 & 80.0 & 76.0 & 47.7 & 72.3 & 77.4 & 71.0 & 31.5 & 61.7 \\
\midrule
 \multirow{5}{*}{2.54}
 & OA-GPTQ & 33.3 & 78.6 & 72.9 & 42.1 & 70.5 & 72.1 & 69.0 & 28.3 & 58.3  \\
    & BSP     & 29.7 & 76.6 & 69.3 & 39.7 & 47.5 & 70.5 & 66.1 & 27.1 & 53.3  \\
    & MxMoE & 29.8 & 79.3 & 72.8 & 43.2 & 67.7 & 74.3 & 68.1 & 29.5 & 58.1  \\
    & PMQ     & 32.8 & 78.9 & 70.1 & 41.8 & 74.2 & 71.0 & 67.9 & 29.3 & 58.2 \\
    & {\cellcolor{charcoal!12}} Colla-Q
   & {\cellcolor{charcoal!12}} 33.4
   & {\cellcolor{charcoal!12}} 79.8
   & {\cellcolor{charcoal!12}} 74.7
   & {\cellcolor{charcoal!12}} 42.8
   & {\cellcolor{charcoal!12}} 72.3
   & {\cellcolor{charcoal!12}} 74.1
   & {\cellcolor{charcoal!12}} 69.1
   & {\cellcolor{charcoal!12}} 29.8
   & {\cellcolor{charcoal!12}} 59.5
    \\
\midrule
 \multirow{3}{*}{2.05}
 & MxMoE & 26.4 & 75.7 & 67.1 & 37.6 & 66.3 & 64.4 & 66.2 & 27.0 & 53.8  \\
  & PMQ  & 27.0 & 77.3 & 64.4 & 38.6 & 67.5 & 68.0 & 65.1 & 25.3 & 54.1 \\
    & {\cellcolor{charcoal!12}} Colla-Q
   & {\cellcolor{charcoal!12}} 27.1
   & {\cellcolor{charcoal!12}} 77.0
   & {\cellcolor{charcoal!12}} 68.7
   & {\cellcolor{charcoal!12}} 38.7
   & {\cellcolor{charcoal!12}} 69.1
   & {\cellcolor{charcoal!12}} 67.3
   & {\cellcolor{charcoal!12}} 66.5
   & {\cellcolor{charcoal!12}} 26.8
   & {\cellcolor{charcoal!12}} 55.2 \\
\midrule
 \multirow{3}{*}{1.57}&  MxMoE & 24.0 & 63.9 & 40.5 & 23.9 & 63.4 & 43.0 & 59.1 & 24.2 & 42.8  \\

  & PMQ  & 22.6 & 73.0 & 62.0 & 34.6 & 63.0 & 58.3 & 63.4 & 23.5 & 50.1  \\
    & {\cellcolor{charcoal!12}} Colla-Q
   & {\cellcolor{charcoal!12}} 23.3
   & {\cellcolor{charcoal!12}} 72.6
   & {\cellcolor{charcoal!12}} 64.3
   & {\cellcolor{charcoal!12}} 35.1
   & {\cellcolor{charcoal!12}} 64.0
   & {\cellcolor{charcoal!12}} 57.6
   & {\cellcolor{charcoal!12}} 65.0
   & {\cellcolor{charcoal!12}} 24.5
   & {\cellcolor{charcoal!12}} 50.8
    \\
\midrule
\bottomrule
\end{tabular}
\caption{Zero-shot task performance of DeepSeek-16B-Base on eight benchmarks under different average bit-widths and quantization methods.}
\label{tab:deepseek_total}
\end{table*}
}

\newcommand{\phitotaltable}{
\begin{table*}[t]
\centering
\small
\setlength{\tabcolsep}{7.7pt}
\renewcommand{\arraystretch}{1}
\definecolor{charcoal}{RGB}{54,69,79}
\begin{tabular}{ccccccccccc}
\toprule
\midrule
\textbf{Bits} & \textbf{Method} & \textbf{MMLU} & \textbf{PIQA} & \textbf{ARC-e} & \textbf{ARC-c} & \textbf{BoolQ} & \textbf{HellaS.} & \textbf{Wino.} & \textbf{MathQA} & \textbf{Avg.}  \\
\midrule
 16 & - & 76.6 & 78.2 & 65.0 & 53.6 & 88.4 & 79.1 & 75.9 & 37.3 & 69.3 \\
\midrule

\multirow{5}{*}{2.54}& OA-GPTQ & 43.8 & 66.0 & 55.7 & 40.1 & 69.9 & 60.9 & 58.3 & 22.6 & 52.2  \\
    & BSP     & 39.4 & 60.0 & 42.4 & 33.8 & 56.5 & 48.3 & 51.5 & 22.0 & 44.2  \\
    & MxMoE & 43.9 & 64.1 & 51.6 & 38.9 & 64.7 & 57.8 & 58.5 & 23.3 & 50.3 \\
    & PMQ     & 50.7 & 71.6 & 58.6 & 42.3 & 76.6 & 70.0 & 62.9 & 22.5 & 56.9  \\
    & {\cellcolor{charcoal!12}} Colla-Q
   & {\cellcolor{charcoal!12}} 51.8
   & {\cellcolor{charcoal!12}} 72.1
   & {\cellcolor{charcoal!12}} 59.2
   & {\cellcolor{charcoal!12}} 41.6
   & {\cellcolor{charcoal!12}} 76.3
   & {\cellcolor{charcoal!12}} 70.7
   & {\cellcolor{charcoal!12}} 63.6
   & {\cellcolor{charcoal!12}} 25.5
   & {\cellcolor{charcoal!12}} 57.6
   \\
\midrule

\multirow{3}{*}{2.05}
& MxMoE & 27.8 & 57.9 & 41.3 & 26.5 & 53.7 & 44.6 & 51.5 & 20.7 & 40.5 \\
  & PMQ  & 25.4 & 64.6 & 50.5 & 36.0 & 59.9 & 60.4 & 60.2 & 22.0 & 47.4 \\
    & {\cellcolor{charcoal!12}} Colla-Q
   & {\cellcolor{charcoal!12}} 28.3
   & {\cellcolor{charcoal!12}} 68.1
   & {\cellcolor{charcoal!12}} 51.0
   & {\cellcolor{charcoal!12}} 37.4
   & {\cellcolor{charcoal!12}} 67.6
   & {\cellcolor{charcoal!12}} 61.4
   & {\cellcolor{charcoal!12}} 60.9
   & {\cellcolor{charcoal!12}} 23.8
   & {\cellcolor{charcoal!12}} 49.8
    \\
\midrule

\multirow{3}{*}{1.57}
& MxMoE & 23.7 & 53.1 & 32.3 & 27.1 & 55.4 & 32.4 & 50.0 & 21.1 & 36.8 \\
 &  PMQ  & 23.5 & 56.8 & 37.9 & 31.4 & 50.6 & 43.4 & 53.1 & 19.4 & 39.5 \\
   & {\cellcolor{charcoal!12}} Colla-Q
   & {\cellcolor{charcoal!12}} 24.3
   & {\cellcolor{charcoal!12}} 59.8
   & {\cellcolor{charcoal!12}} 40.8
   & {\cellcolor{charcoal!12}} 29.6
   & {\cellcolor{charcoal!12}} 54.6
   & {\cellcolor{charcoal!12}} 42.7
   & {\cellcolor{charcoal!12}} 54.9
   & {\cellcolor{charcoal!12}} 20.7
   & {\cellcolor{charcoal!12}} 40.9
   \\
\midrule
\bottomrule
\end{tabular}
\caption{Zero-shot task performance of Phi3.5-MoE on eight benchmarks under different average bit-widths and quantization methods.}
\label{tab:phi_total}
\end{table*}
}

\newcommand{\TabModelInfo}{
\begin{table}[t]
\small
\setlength{\tabcolsep}{2.3pt}
\renewcommand{\arraystretch}{1.1}
\centering
\begin{tabular}{lccccc}
\toprule
\midrule
\multirow{2}{*}{\textbf{Model}} & \multirow{2}{*}{\textbf{Layer}} & \multirow{2}{*}{\textbf{\#E}} & \multirow{2}{*}{\textbf{Top-$k$}} & \textbf{Param.} & \textbf{Mem.} \\
& & & & (B) & (GB) \\
\midrule
Mixtral 8$\times$7B & 32 & 8 & 2 & 46.7 & 96.8 \\
DeepSeek-16B-Base & 28 & 64+2 & 6 & 16 & 30.4 \\
Phi3.5-MoE & 32 & 16 & 2 & 42 & 83.8 \\
\midrule
\bottomrule
\end{tabular}
\caption{Architectural and memory configurations of MoE models. \#E denotes the number of experts in each model, and Top-k indicates the number of routed experts in a MoE block.}
\label{tab:model-info}
\end{table}
}

\newcommand{\totalalgorithmtable}{
\begin{table*}[!t]
\centering
\small
\setlength{\tabcolsep}{7.5pt}
\renewcommand{\arraystretch}{1.2}
\definecolor{charcoal}{RGB}{54,69,79}
\begin{tabular}{cccccccccccc}
\toprule
\midrule
\textbf{Bits} & \textbf{Algorithm} & \textbf{MMLU} & \textbf{PIQA} & \textbf{ARC-e} & \textbf{ARC-c} & \textbf{BoolQ} & \textbf{HellaS.} & \textbf{Wino.} & \textbf{MathQA} & \textbf{Avg.} \\
\midrule
\multirow{2}{*}{2.54}
 & PMQ  & 57.6 & 79.5 & 78.4 & 51.9 & 83.5 & 78.8 & 72.4 & 36.4 & 67.3 \\
   & {\cellcolor{charcoal!12}}Colla-Q & {\cellcolor{charcoal!12}}59.4 & {\cellcolor{charcoal!12}}80.5 & {\cellcolor{charcoal!12}}79.8 & {\cellcolor{charcoal!12}}54.0 & {\cellcolor{charcoal!12}}84.7 & {\cellcolor{charcoal!12}}78.5 & {\cellcolor{charcoal!12}}74.9 & {\cellcolor{charcoal!12}}36.5 & {\cellcolor{charcoal!12}}68.5
   \\
\midrule
\multirow{2}{*}{2.05}
 & PMQ  & 49.3 & 77.4 & 72.7 & 45.2 & 81.7 & 73.3 & 71.5 & 33.9 & 63.1 \\
   & {\cellcolor{charcoal!12}}Colla-Q & {\cellcolor{charcoal!12}}52.9 & {\cellcolor{charcoal!12}}77.8 & {\cellcolor{charcoal!12}}74.9 & {\cellcolor{charcoal!12}}46.4 & {\cellcolor{charcoal!12}}83.3 & {\cellcolor{charcoal!12}}72.4 & {\cellcolor{charcoal!12}}72.0 & {\cellcolor{charcoal!12}}32.3 & {\cellcolor{charcoal!12}}64.0  \\
\midrule
\multirow{2}{*}{1.57}
 & PMQ  & 34.2 & 72.1 & 59.1 & 36.2 & 72.3 & 61.3 & 63.4 & 27.2 & 53.2 \\
   & {\cellcolor{charcoal!12}}Colla-Q & {\cellcolor{charcoal!12}}36.6 & {\cellcolor{charcoal!12}}71.0 & {\cellcolor{charcoal!12}}63.1 & {\cellcolor{charcoal!12}}36.2 & {\cellcolor{charcoal!12}}75.8 & {\cellcolor{charcoal!12}}59.8 & {\cellcolor{charcoal!12}}66.2 & {\cellcolor{charcoal!12}}26.9 & {\cellcolor{charcoal!12}}54.5  \\
\midrule
\bottomrule
\end{tabular}
\caption{Zero-shot performance of Mixtral 8$\times$7B with different bit-width allocation algorithms. We compare PMQ's allocation algorithm and our minimax precision balancing using the proposed activation-entropy metric.}
\label{tab:totalalrorithmtable}
\end{table*}}

\newcommand{\calibdatasetablationcombined}{
\begin{table*}[!ht]
\centering
\small
\setlength{\tabcolsep}{4.5pt}
\renewcommand{\arraystretch}{1.}
\begin{tabular}{cccccccccc}
\toprule
\midrule
 \multirow{2}{*}{\textbf{Method}} &
\multirow{2}{*}{\textbf{\shortstack{Calibration\\dataset}}} & \multicolumn{4}{c}{\textbf{Mixtral 8$\times$7B}} & \multicolumn{4}{c}{\textbf{Phi3.5-MoE}} \\
\cmidrule(lr){3-6}\cmidrule(lr){7-10}

& & \textbf{MathQA} & \textbf{Lambada\_mul\_fr} & \textbf{BoolQ} & \textbf{Avg.} &
\textbf{MathQA} & \textbf{Lambada\_mul\_fr} & \textbf{BoolQ} & \textbf{Avg.}  \\
\midrule
\multirow{4}{*}{PMQ}
 & C4     & 26.8 & 23.8 & 67.8 & 39.4 & 19.4 & 0.8 & 50.6 & 23.6 \\
 & Math   & 27.0 & 22.9 & 68.2 & 39.4 & 22.1 & 0.1 & 54.0 & 25.4 \\
 & French & 25.4 & 25.7 & 72.1 & 41.1 & 19.3 & 4.4 & 51.6 & 25.1 \\
 & QA     & 25.8 & 21.3 & 76.2 & 41.1 & 20.3 & 0.6 & 57.3 & 26.1 \\
\midrule
\multirow{4}{*}{Ours}
 & C4     & 26.9 & 26.8 & 76.5 & 43.5 & 22.7 & 6.8 & 60.6 & 30.0 \\
 & Math   & 27.3 & 26.4 & 77.8 & 43.8 & 23.0 & 6.3 & 60.8 & 30.0 \\
 & French & 27.2 & 26.9 & 77.8 & 44.0 & 22.1 & 6.6 & 61.3 & 30.0 \\
 & QA     & 27.1 & 26.3 & 77.6 & 43.7 & 22.2 & 6.5 & 61.5 & 30.0 \\
 \midrule
\bottomrule
\end{tabular}
\caption{Zero-shot task performance under different calibration datasets (C4, Math, French, QA), comparing PMQ and ours for Mixtral 8$\times$7B and Phi3.5-MoE in the 1.57-bit setting.}
\label{tab:calb_combined}
\end{table*}
}

\newcommand{\linearexpertcomparefull}{
\begin{table*}[t]
\centering
\small
\setlength{\tabcolsep}{7.2pt}
\renewcommand{\arraystretch}{1.2}
\begin{tabular}{ccccccccccc}
\toprule
\midrule
\textbf{Bits} & \textbf{Components} & \textbf{MMLU} & \textbf{PIQA} & \textbf{ARC-e} & \textbf{ARC-c} & \textbf{BoolQ} & \textbf{HellaS.} & \textbf{Wino.} & \textbf{MathQA} & \textbf{Avg.} \\
\midrule
16  & - & 67.8 & 83.6 & 84.2 & 56.5 & 85.0 & 84.0 & 76.2 & 41.7 & 72.4  \\
\midrule
\multirow{2}{*}{2.54} & Linear Layer & 57.9 & 80.6 & 80.7 & 53.4 & 84.9 & 78.8 & 74.7 & 36.4 & 68.4  \\
 & {\cellcolor{charcoal!12}}Expert &
 {\cellcolor{charcoal!12}}59.4 & {\cellcolor{charcoal!12}}80.5 & {\cellcolor{charcoal!12}}79.8 & {\cellcolor{charcoal!12}}54.0 & {\cellcolor{charcoal!12}}84.7 & {\cellcolor{charcoal!12}}78.5 & {\cellcolor{charcoal!12}}74.9 & {\cellcolor{charcoal!12}}36.5 & {\cellcolor{charcoal!12}}68.5  \\
 \midrule
\multirow{2}{*}{2.05} & Linear Layer & 49.1 & 78.3 & 72.2 & 45.2 & 81.7 & 72.8 & 72.8 & 33.0 & 63.1  \\
 & {\cellcolor{charcoal!12}}Expert & {\cellcolor{charcoal!12}}52.9 & {\cellcolor{charcoal!12}}77.8 & {\cellcolor{charcoal!12}}74.9 & {\cellcolor{charcoal!12}}46.4 & {\cellcolor{charcoal!12}}83.3 & {\cellcolor{charcoal!12}}72.4 & {\cellcolor{charcoal!12}}72.0 & {\cellcolor{charcoal!12}}32.3 & {\cellcolor{charcoal!12}}64.0  \\
 \midrule
\multirow{2}{*}{1.57} & Linear Layer & 34.1 & 71.1 & 61.8 & 34.5 & 63.1 & 56.6 & 65.8 & 27.2 & 51.8  \\
 & {\cellcolor{charcoal!12}}Expert &{\cellcolor{charcoal!12}}36.6 & {\cellcolor{charcoal!12}}71.0 & {\cellcolor{charcoal!12}}63.1 & {\cellcolor{charcoal!12}}36.2 & {\cellcolor{charcoal!12}}75.8 & {\cellcolor{charcoal!12}}59.8 & {\cellcolor{charcoal!12}}66.2 & {\cellcolor{charcoal!12}}26.9 & {\cellcolor{charcoal!12}}54.5 \\
 \midrule
 \bottomrule
\end{tabular}
\caption{Zero-shot task performance of Mixtral 8$\times$7B on eight benchmarks, comparing mixed-precision bit-width allocation across components (Linear Layer vs. Expert) under different average bit budgets.}
\label{tab:linearexpertfull}
\end{table*}}

\newcommand{\gatingweight}{
\begin{table}[ht]
\centering
\small
\setlength{\tabcolsep}{3.5pt}
\renewcommand{\arraystretch}{1}
\begin{tabular}{ccccccccc}
\toprule
\midrule
\textbf{Metric} & \textbf{E0} & \textbf{E1} & \textbf{E2} & \textbf{E3} & \textbf{E4} & \textbf{E5} & \textbf{E6} & \textbf{E7} \\
\midrule
Avg. & 0.50 & 0.51 & 0.50 & 0.49 & 0.50 & 0.50 & 0.51 & 0.49 \\
\midrule
\bottomrule
\end{tabular}
\caption{Conditional average routing weight of each expert over many tokens, averaged over Layers 00-31.}
\label{tab:gatingweight}
\end{table}}

\newcommand{\gatingweightmad}{
\begin{table}[ht]
\centering
\small
\setlength{\tabcolsep}{2.5pt}
\renewcommand{\arraystretch}{1}
\begin{tabular}{cc|cc|cc|cc}
\toprule
\midrule
\textbf{Layer} & \textbf{MAD}  & \textbf{Layer} & \textbf{MAD}  & \textbf{Layer} & \textbf{MAD}  & \textbf{Layer} & \textbf{MAD}  \\
\midrule
0     & 0.06 & 8     & 0.01 & 16    & 0.03 & 24    & 0.02  \\
1     & 0.07 & 9     & 0.02 & 17    & 0.03 & 25    & 0.04  \\
2     & 0.02 & 10    & 0.03 & 18    & 0.02 & 26    & 0.04  \\
3     & 0.01 & 11    & 0.03 & 19    & 0.02 & 27    & 0.04  \\
4     & 0.01 & 12    & 0.04 & 20    & 0.02 & 28    & 0.06  \\
5     & 0.01 & 13    & 0.01 & 21    & 0.01 & 29    & 0.04  \\
6     & 0.02 & 14    & 0.02 & 22    & 0.02 & 30    & 0.06  \\
7     & 0.01 & 15    & 0.01 & 23    & 0.02 & 31    & 0.08  \\
\midrule
\bottomrule
\end{tabular}
\caption{Layer-wise MAD of the conditional average routing weights from the reference value of 0.5.}
\label{tab:gatingweightmad}
\end{table}}

\newcommand{\gatingweightmadall}{
\begin{table}[ht]
\centering
\small
\setlength{\tabcolsep}{0.8pt}
\renewcommand{\arraystretch}{1}
\begin{tabular}{ccccc}
\toprule
\midrule
\multirow{2}{*}{Model}
& \#Experts
& Reference
& Expert-
& Layer-wise \\
& / Top-$k$
& value
& wise avg.
& MAD avg. \\
\midrule
\begin{tabular}[l]{@{}c@{}}
DeepSeek-MoE\\
16B-Base
\end{tabular}
    & 64 / 6
    & $1/6 \approx 0.167$
    & 0.165
    & 0.024 \\
Phi3.5-MoE
    & 16 / 2
    & $1/2 = 0.500$
    & 0.496
    & 0.030 \\
\midrule
\bottomrule
\end{tabular}
\caption{
Conditional average routing weights for additional MoE architectures.
Expert-wise avg. is the mean routing weight; Layer-wise MAD avg. is the mean absolute deviation from the reference value.}
\label{tab:gatingweightmadall}
\end{table}}

\newcommand{\gaussianquantifyall}{
\begin{table}[ht]
\centering
\small
\setlength{\tabcolsep}{2pt}
\renewcommand{\arraystretch}{1.1}
\begin{tabular}{lcc}
\toprule
\midrule
\textbf{Metric} & \textbf{Channel-wise} & \textbf{Expert-wise} \\
\midrule
\multicolumn{3}{c}{\textbf{Mixtral 8$\times$7B}} \\
\midrule
Number of units
    & 32,768 channels
    & 256 experts \\
Mean $L_1$ to $\mathcal{N}(0,1)$
    & 0.020
    & 0.030 \\
Median $L_1$ to $\mathcal{N}(0,1)$
    & 0.020
    & 0.020 \\
Fraction with $L_1 < 0.05$
    & 0.99
    & 0.92 \\
\midrule
\multicolumn{3}{c}{\textbf{DeepSeek-MoE-16B-Base}} \\
\midrule
Number of units
    & 228,096 channels
    & 1,782 experts \\
Mean $L_1$ to $\mathcal{N}(0,1)$
    & 0.018
    & 0.015 \\
Median $L_1$ to $\mathcal{N}(0,1)$
    & 0.016
    & 0.012 \\
Fraction with $L_1 < 0.05$
    & 0.98
    & 0.99 \\
\midrule
\multicolumn{3}{c}{\textbf{Phi3.5-MoE}} \\
\midrule
Number of units
    & 65,536 channels
    & 512 experts \\
Mean $L_1$ to $\mathcal{N}(0,1)$
    & 0.0256
    & 0.029 \\
Median $L_1$ to $\mathcal{N}(0,1)$
    & 0.022
    & 0.023 \\
Fraction with $L_1 < 0.05$
    & 0.95
    & 0.90 \\
\midrule
\bottomrule
\end{tabular}
\caption{Quantitative results for the Gaussianity of channel-wise and expert-wise expert FFN output activations. Lower L1 values indicate that the distribution is closer to the standard Gaussian distribution. Fraction with L1 $< 0.05$ denotes the proportion of channels or experts with L1 distance below 0.05.}
\label{tab:gaussianquantify}
\end{table}
}

\newcommand{\algorithmmetirctable}{
\begin{table}[t]
\centering
\small
\setlength{\tabcolsep}{12pt}
\renewcommand{\arraystretch}{1.1}
\definecolor{charcoal}{RGB}{54,69,79}
\begin{tabular}{cccc}
\toprule
\midrule
\multirow{2}{*}{\textbf{Bits}} &
\multirow{2}{*}{\textbf{Metric}} &
\multicolumn{2}{c}{\textbf{Algorithm}} \\
\cmidrule(lr){3-4}
&
&
ILP &
Ours \\
\midrule

\multirow{2}{*}{2.54} & PMQ  & 67.5 & 67.7 \\
     & Ours & 67.3 & {\cellcolor{charcoal!12}}68.5 \\

\multirow{2}{*}{2.05} & PMQ  & 63.3 & 63.3 \\
     & Ours & 63.1 & {\cellcolor{charcoal!12}}64.0 \\

\multirow{2}{*}{1.57} & PMQ  & 54.5 & 53.6 \\
     & Ours & 53.2 & {\cellcolor{charcoal!12}}54.5 \\

\midrule
\bottomrule
\end{tabular}
\caption{Ablation study on Mixtral 8$\times$7B comparing combinations of PMQ and activation entropy metrics with ILP and minimax precision balancing. Values report average performance across eight benchmarks.}
\label{tab:algorithmmetirctable}
\end{table}}

\newcommand{\quantlosstable}{
\begin{table}
\centering
\small
\setlength{\tabcolsep}{6pt}
\renewcommand{\arraystretch}{1.05}
\definecolor{charcoal}{RGB}{54,69,79}
\begin{tabular}{ccccc}
\toprule
\midrule
\textbf{Bits} & \textbf{Term} & \textbf{MMLU} & \textbf{C.S Avg.} & \textbf{Avg.} \\
\midrule

16 & - & 67.8 & 73.0 & 70.4 \\

\midrule
\multirow{2}{*}{2.54}
& Quantization error & 58.3 & 69.2 & 63.7 \\
& {\cellcolor{charcoal!12}}Colla-Q
& {\cellcolor{charcoal!12}}59.4
& {\cellcolor{charcoal!12}}69.8
& {\cellcolor{charcoal!12}}64.6 \\

\midrule
\multirow{2}{*}{2.05}
& Quantization error & 49.9 & 65.8 & 57.8 \\
& {\cellcolor{charcoal!12}}Colla-Q
& {\cellcolor{charcoal!12}}52.9
& {\cellcolor{charcoal!12}}65.6
& {\cellcolor{charcoal!12}}59.3 \\

\midrule
\multirow{2}{*}{1.57}
& Quantization error & 35.7 & 56.0 & 45.8 \\
& {\cellcolor{charcoal!12}}Colla-Q
& {\cellcolor{charcoal!12}}36.6
& {\cellcolor{charcoal!12}}57.0
& {\cellcolor{charcoal!12}}46.8 \\

\midrule
\bottomrule
\end{tabular}
\caption{Zero-shot performance of Mixtral 8$\times$7B under different metric settings. We compare quantization error with activation-entropy-weighted quantization error using the same minimax precision balancing algorithm.}
\label{tab:quantlosstable}
\end{table}}

\newcommand{\dilpargo}{
\begin{table}[t]
\centering
\small
\setlength{\tabcolsep}{6pt}
\renewcommand{\arraystretch}{1.1}
\definecolor{charcoal}{RGB}{54,69,79}
\begin{tabular}{cccc}
\toprule
\midrule
\textbf{Bits} &
\shortstack{\textbf{Minimax Precision}\\\textbf{Balancing (Ours)}} &
\shortstack{\textbf{Direct ILP}\\\textbf{w/ Colla-Q Objective}} \\
\midrule

2.54 & {\cellcolor{charcoal!12}}68.5 & 65.9 \\
2.05 & {\cellcolor{charcoal!12}}64.0 & 61.7 \\
1.57 & {\cellcolor{charcoal!12}}54.5 & 52.9 \\

\midrule
\bottomrule
\end{tabular}
\caption{Ablation study of average zero-shot performance on eight benchmarks for Mixtral 8$\times$7B. We compare minimax precision balancing with Direct ILP using the same Colla-Q objective and bit budget.}
\label{tab:directilp}
\end{table}}

\newcommand{\quantlosscomparefull}{
\begin{table*}[t]
\centering
\small
\setlength{\tabcolsep}{6.5pt}
\renewcommand{\arraystretch}{1.2}
\begin{tabular}{ccccccccccc}
\toprule
\midrule
\textbf{Bits} & \textbf{Term} & \textbf{MMLU} & \textbf{PIQA} & \textbf{ARC-e} & \textbf{ARC-c} & \textbf{BoolQ} & \textbf{HellaS.} & \textbf{Wino.} & \textbf{MathQA} & \textbf{Avg.} \\
\midrule
16  & - & 67.8 & 83.6 & 84.2 & 56.5 & 85.0 & 84.0 & 76.2 & 41.7 & 72.4 \\
\midrule
\multirow{2}{*}{2.54} & Quantization error & 58.3 & 80.0 & 79.0 & 51.4 & 83.9 & 79.0 & 75.0 & 35.8 & 67.8 \\
 & {\cellcolor{charcoal!12}}Colla-Q &
 {\cellcolor{charcoal!12}}59.4 & {\cellcolor{charcoal!12}}80.5 & {\cellcolor{charcoal!12}}79.8 & {\cellcolor{charcoal!12}}54.0 & {\cellcolor{charcoal!12}}84.7 & {\cellcolor{charcoal!12}}78.5 & {\cellcolor{charcoal!12}}74.9 & {\cellcolor{charcoal!12}}36.5 & {\cellcolor{charcoal!12}}68.5 \\
\midrule
\multirow{2}{*}{2.05} & Quantization error & 49.9 & 78.2 & 73.6 & 47.8 & 83.0 & 73.2 & 72.1 & 32.7 & 63.8 \\
 & {\cellcolor{charcoal!12}}Colla-Q &
 {\cellcolor{charcoal!12}}52.9 & {\cellcolor{charcoal!12}}77.8 & {\cellcolor{charcoal!12}}74.9 & {\cellcolor{charcoal!12}}46.4 & {\cellcolor{charcoal!12}}83.3 & {\cellcolor{charcoal!12}}72.4 & {\cellcolor{charcoal!12}}72.0 & {\cellcolor{charcoal!12}}32.3 & {\cellcolor{charcoal!12}}64.0 \\
\midrule
\multirow{2}{*}{1.57} & Quantization error & 35.7 & 70.3 & 63.0 & 34.0 & 76.0 & 58.1 & 64.1 & 26.2 & 53.4 \\
 & {\cellcolor{charcoal!12}}Colla-Q &
 {\cellcolor{charcoal!12}}36.6 & {\cellcolor{charcoal!12}}71.0 & {\cellcolor{charcoal!12}}63.1 & {\cellcolor{charcoal!12}}36.2 & {\cellcolor{charcoal!12}}75.8 & {\cellcolor{charcoal!12}}59.8 & {\cellcolor{charcoal!12}}66.2 & {\cellcolor{charcoal!12}}26.9 & {\cellcolor{charcoal!12}}54.5 \\
\midrule
\bottomrule
\end{tabular}
\caption{Zero-shot task performance of Mixtral 8$\times$7B on eight benchmarks, comparing quantization error with Colla-Q under different average bit budgets.}
\label{tab:quantlossfull}
\end{table*}}

\newcommand{\expertpair}{
\begin{table}[t]
\centering
\small
\setlength{\tabcolsep}{6pt}
\renewcommand{\arraystretch}{1.1}
\definecolor{charcoal}{RGB}{54,69,79}
\begin{tabular}{cccc}
\toprule
\midrule
\textbf{Layer} & \textbf{Spearman Correlation} \\
\midrule
1  & $-0.67$ \\
7  & $-0.57$ \\
11 & $-0.43$ \\
\midrule
{\cellcolor{charcoal!12}}Average & {\cellcolor{charcoal!12}}$-0.56$ \\
\midrule
\bottomrule
\end{tabular}
\caption{Spearman correlation between activation entropy $\rho$ and
expert-level performance obtained from the expert-pair intervention.
A negative correlation indicates that lower $\rho$ tends to correspond
to higher expert-level performance.}
\label{tab:expert_pair}
\end{table}}